\documentclass[10pt, conference, letterpaper]{IEEEtran}
\usepackage{amsmath,dsfont,mathtools}
\usepackage{tabularx,multirow}
\usepackage{graphicx,subfigure,comment}
\usepackage[table]{xcolor}
\usepackage{url} 
\usepackage{dsfont}
\usepackage{cite}
\usepackage{pstool}
\usepackage{enumitem,amsmath}
\usepackage{algorithm}
\usepackage{algpseudocodex}

\newcommand{\Fig}[1]{Fig.~\ref{fig:#1}}
\newcommand{\Prop}[1]{Property~\ref{prop:#1}}

\newcommand{\Sec}[1]{Sec.~\ref{sec:#1}}
\newcommand{\Tab}[1]{Tab.~\ref{tab:#1}}
\newcommand{\Eq}[1]{(\ref{eq:#1})}
\newcommand{\Alg}[1]{Alg.~\ref{alg:#1}}
\newcommand{\Line}[1]{Line~\ref{line:#1}}

\newcommand{\Mc}{\mathcal{M}}
\newcommand{\Nc}{\mathcal{N}}
\newcommand{\Pc}{\mathcal{P}}
\newcommand{\Gc}{\mathcal{G}}
\newcommand{\Vc}{\mathcal{V}}
\newcommand{\Ec}{\mathcal{E}}
\newcommand{\Tc}{\mathcal{T}}
\newcommand{\Wc}{\mathcal{W}}

\newcommand{\er}{\mathrm{e}}
\newcommand{\EE}{\mathds{E}}

\newcommand{\Ab}{\mathbf{A}}
\newcommand{\Cb}{\mathbf{C}}
\newcommand{\Vb}{\mathbf{V}}
\newcommand{\ab}{\mathbf{a}}
\renewcommand{\sb}{\mathbf{s}}

\newtheorem{property}{Property}

\DeclareMathOperator\erfc{erfc}
\DeclarePairedDelimiter\ceil{\lceil}{\rceil}

\begin{document}

\title{Matching DNN Compression and Cooperative Training  
with Resources and Data Availability}

\author{
F.~Malandrino$^{1,2}$, G.~Di~Giacomo$^{3}$, A.~Karamzade$^{4}$, M.~Levorato$^{4}$, C.~F.~Chiasserini$^{3,1,2}$\\
1: CNR-IEIIT, Italy -- 2: CNIT, Italy -- 3: Politecnico di Torino, Italy -- 4: UC Irvine, USA
\vspace{-1 cm}
}

\maketitle

\begin{abstract}
To make machine learning (ML) sustainable and apt to run on the diverse devices where relevant data is, it is essential to compress ML models as needed, while still meeting the required learning quality and time performance. However, {\em how much} and {\em when} an ML model should be compressed, and {\em where} its training should be executed, are hard decisions to make, as they depend on the model itself, the resources of the available nodes, and the data such nodes own. 
Existing studies  focus on each of those aspects individually, however, they do not account for how such decisions can be made jointly and adapted to one another. In this work, 
we  model the network system focusing on the training of DNNs, formalize the above  multi-dimensional problem, and, given its NP-hardness, formulate an approximate dynamic programming problem that we solve through the PACT algorithmic framework. Importantly, PACT leverages a time-expanded graph representing the learning process, and a data-driven  and theoretical approach for the prediction of the loss evolution to be expected as a consequence of training decisions. We prove that PACT’s solutions can get as close to the optimum as desired, at the cost of an increased time complexity, and that, in any case, such  complexity is polynomial. Numerical results also show that, even under the most disadvantageous settings, PACT outperforms state-of-the-art alternatives and closely matches the optimal energy cost.
\end{abstract}

\section{Introduction}
\label{sec:intro}

Modern-day machine-learning (ML) models are hard to train, as they often require considerable quantities of data as well as computational, network and energy resources~\cite{callegaro2021optimal,tehrani2021federated}. In addition, in several application scenarios, data and resources may be located across different network nodes, whose availability and connectivity may significantly differ, and even vary in space and time~\cite{callegaro2022smartdet}. Examples include smart factory ML-based applications, where models are partially trained in the cloud and then specialized by edge nodes \cite{smart-factory}, or even more extreme settings where image classification models are first trained by ground stations and then refined by a spacecraft specifically for its operating environment \cite{space}. 

In all the above cases, a critical {\em technical challenge} is the {\em mutual adaptation of the training process and the system settings, resources and data} offered by the interconnected network nodes. We contend that existing works only address some specific aspects but none of them tackles the joint optimization of ML model compression and the selection of nodes and related data.  
The framework we propose achieves such goals by leveraging {\em multiple} ML models throughout different stages of a single learning task -- {\em switching} among them as needed  (e.g., via  model pruning~\cite{wen2016learning}, or knowledge distillation (KD)~\cite{gou2021knowledge}) -- and choosing, for each stage, the most appropriate datasets and resources. To make an example, the training of a complex model can start over a small set of powerful nodes; then, we can switch to a simpler, compressed model  and perform further training epochs involving additional nodes that contribute fewer resources but more valuable data \cite{smart-factory2} toward their specific task and domain. Model and nodes switching, however, comes with its own costs in terms of time, resources and, often, learning performance, hence, switching decisions must be made only when clear gains can be obtained.

In this work, we formalize and optimize such complex cooperative strategies for the training of deep neural networks (DNNs), where the -- heterogeneous -- nodes that participate in the training do not share their data. The training process is illustrated in Fig.~\ref{fig:pact}.
In this challenging scenario, we make the following contributions:

\begin{figure}
\centering
\includegraphics[width=\columnwidth]{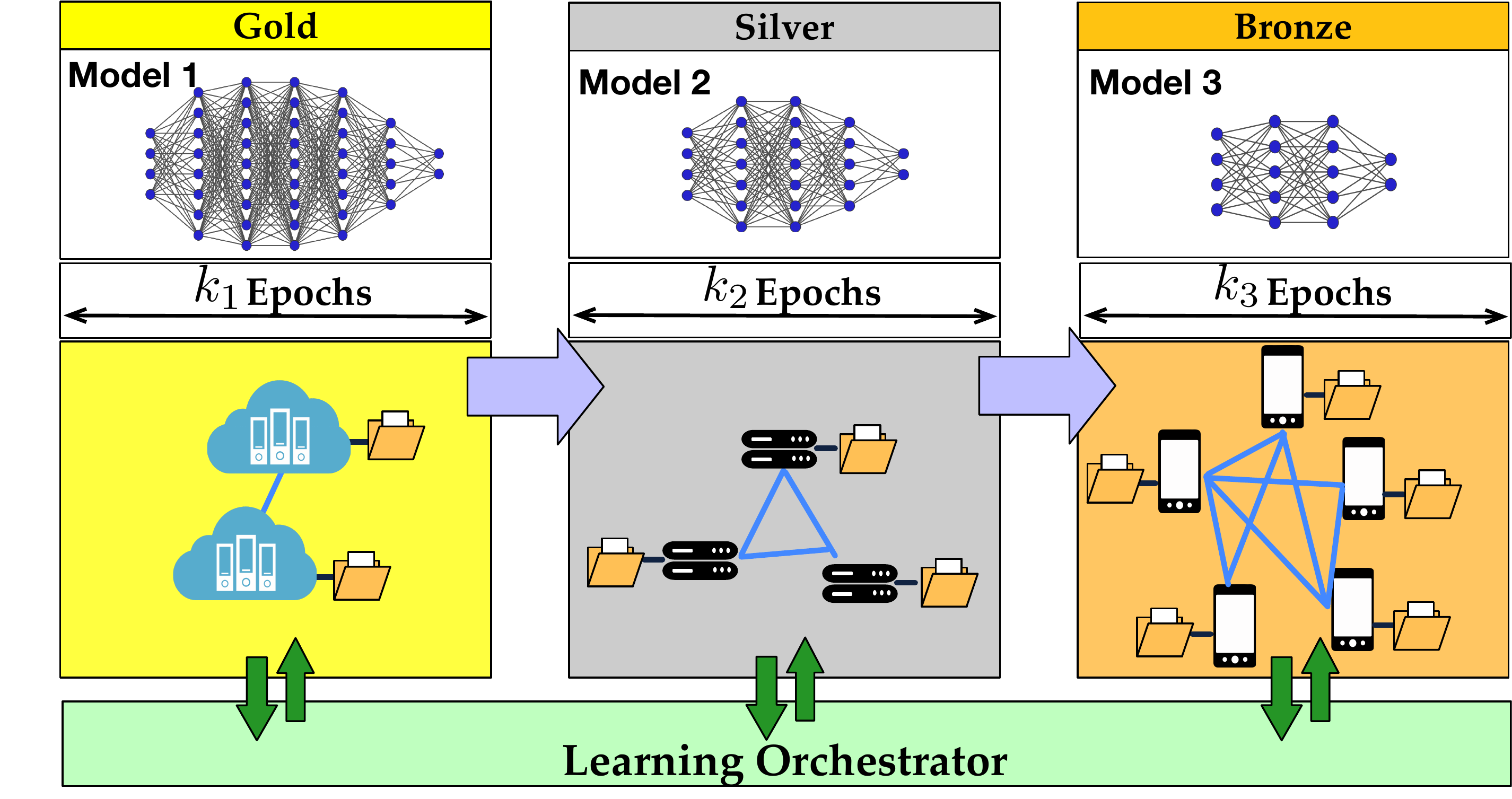}\vspace{-3mm}
\caption{Cooperative training process proposed and optimized in this paper. Subsets of nodes sequentially train compressed versions of an original DNN model. 
In the picture, we categorize nodes based on their computing capabilities and data availability, and, in the example, the training sequence is based on nodes' ranking (gold, silver, bronze). Our framework, named PACT and running at the learning orchestrator, can generate arbitrary sequences: it optimizes the set of nodes, number of epochs, and model compression along the process.  \label{fig:pact}\vspace{-4mm}}
\end{figure}

\noindent
(1) {\em Model and problem definition:} We develop a model for the networked system supporting the training of DNN models  capturing its most relevant aspects, and we use it to make {\em dynamic, joint} decisions about: (i) the ML {\em model} to be used at each epoch (e.g., the full DNN model or a compressed version thereof); (ii) {\em when} to perform a model switch (e.g., at which epoch the framework transitions to a compressed version of the DNN); and (iii) the {\em nodes} to leverage at each stage, accounting for their resources and specific local datasets. The overarching goal is to reach a target learning quality 
by a desired deadline, while minimizing the energy consumption (hence, the cost) of the overall process. 
We remark that controlling {\em all} the above aspects allows for more flexibility  than existing works that just study how to adapt one aspect (e.g., choosing the model) to another (e.g., a given and immutable set of resources). 

\noindent
(2) {\em Algorithmic framework:} The decision process is further complicated by two main issues, namely, (i) the scale of the problem itself, and (ii) the fact that the effect of model switching decisions cannot be, in general, known with certainty. We tackle the first issue by adopting an approximate dynamic programming (ADP) approach, whereby only the most promising courses of action are evaluated, and we envision an algorithmic solution, named Performance-Aware Compression and Training (PACT), for an efficient selection thereof. Concerning the second, we integrate into our decision-making approach the available results on performance characterization of DNN model training. In particular, we leverage both theoretical bounds and data-driven predictions based on low-complexity NN architectures of the loss evolution to be expected as a consequence of a sequence of training decisions. 
 Importantly, we prove that PACT  can get as close as desired to the optimum  of the formulated problem (which is shown to be NP-hard), at the cost of an increased complexity, which is anyway polynomial at worst. 

\noindent
(3) {\em Performance evaluation:} 
We  show how PACT identifies the  training strategy that best matches the available resources and data, resulting in minimum energy consumption given the target loss value and training process time. Also, PACT demonstrates to be highly robust to  approximate estimations of the effects of model switching on the loss values.

In the rest of the paper,  \Sec{motivating} clarifies 
the problem we address, while \Sec{model} presents the system model and  the decisions we tackle. 
\Sec{estimating} then introduces the methodology used for estimating the loss as learning proceeds, and \Sec{buddy} describes our algorithmic solution. The obtained results are shown in \Sec{results}; finally, \Sec{relwork} discusses relevant related work and \Sec{concl} summarizes our conclusions.

\section{A Motivating Example
\label{sec:motivating}}

In this section, we illustrate the benefits of a cooperative training process that integrates model and nodes switching, but also emphasize the challenges in formulating and optimizing it. To this aim, we consider  the case in which one of the most popular cooperative learning approaches, namely, federated learning (FL), is coupled with model pruning~\cite{wen2016learning}. The latter exploits the fact that, typically, many of a model's parameters have a small impact on its performance and can thus be pruned away, resulting in a DNN with similar performance  but of lower complexity, and hence CPU and memory requirements. In particular,  we evaluate the following scenario:
\begin{itemize}
    \item the nodes perform an image classification task using the  
    VGG11 DNN model \cite{simonyan2014very} as a starting point;
    \item FL uses the cross entropy loss function, batch size equal to 64, and the gradient descent optimizer with $10^{-3}$ learning rate  and 0.9 momentum;
    
    \item the model is trained for~$K_1$ epochs on 5~highly capable nodes (``gold'' nodes), each using 8,000~randomly-chosen images from the CIFAR-10 dataset~\cite{krizhevsky2009learning};
    \item then, a fraction~$F$ of the model's parameters is pruned
    \item finally, training resumes adding 2 more learning nodes, which have  lower computing capability and fewer data: either ``silver'' with half the computing resources of the gold nodes and 2,500 local images each, or ``bronze'' with one third of the computing resources of the gold nodes and 750 local images each.
\end{itemize}

\begin{figure}
\centering
\includegraphics[width=\columnwidth]{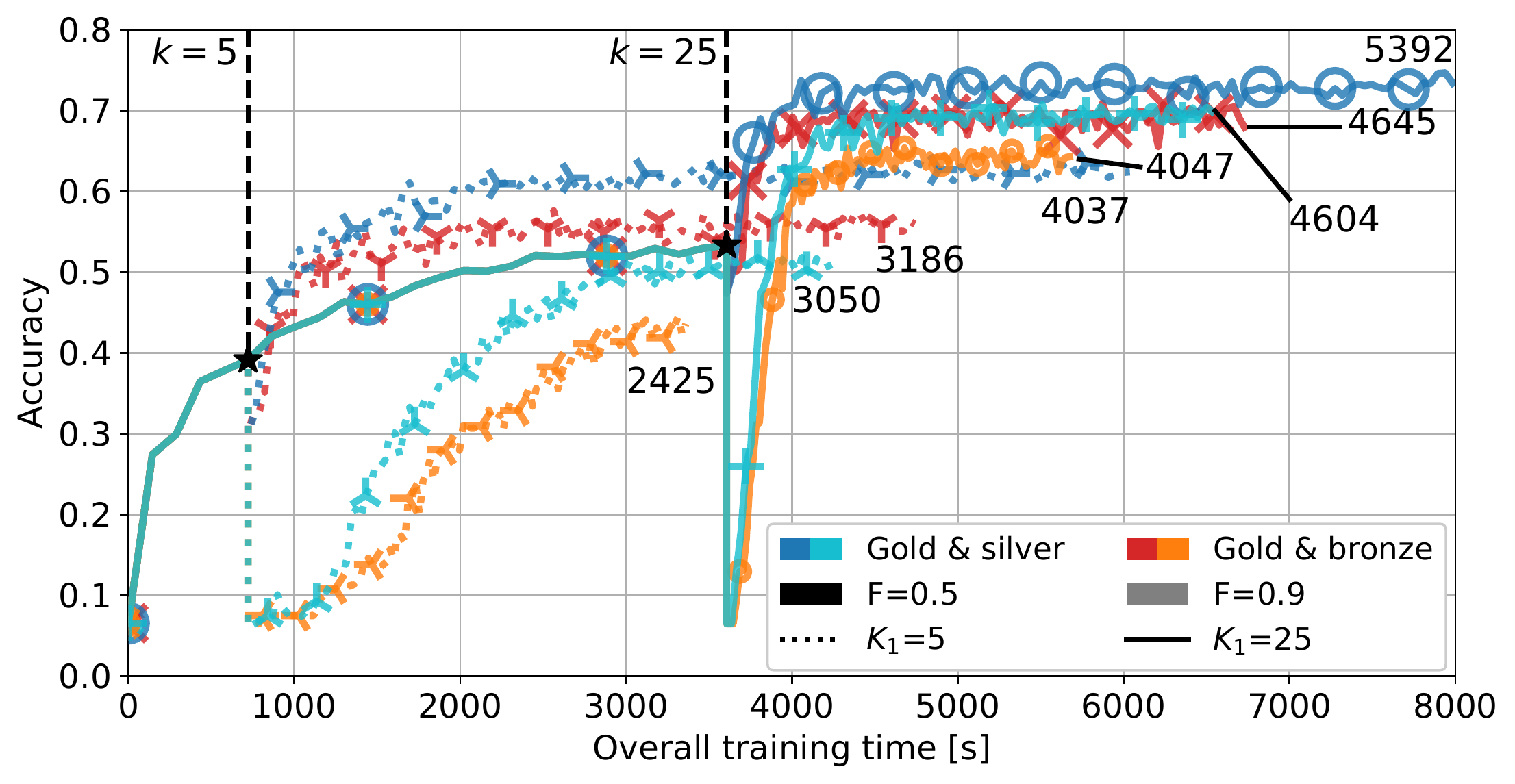}\vspace{-4mm}
\caption{Accuracy vs. training time for different values 
of pruning epoch~$K_1$ (denoted by different line styles) and percentage~$F$ 
(denoted by different color shades). Note how upon pruning a sudden drop in accuracy occurs. Cold and warm colors denote the set of nodes used 
for FL. Numbers in the plot indicate the total CPU time [s], while each marker corresponds to 10  epochs.\label{fig:pruning}\vspace{-4mm}}
\end{figure}

Three decisions should be made: (i) the number~$K_1$ of epochs to execute before pruning, (ii) the percentage~$F$ of 
parameters to prune, and (iii) whether to use the ``silver'' or ``bronze'' nodes when resuming training. Notice how the first two decisions 
concern selecting and switching among models, while the third deals with the physical nodes participating in the learning process. 
\Fig{pruning} summarizes the effects of such decisions\footnote{Only some values of~$F$ are possible, as we apply structured pruning (see \Sec{relwork} for further details).}, which lead to  the following main remarks.\\
{\em Observation 1:} Pruning more (i.e., $F=0.9$, orange and light blue curves) significantly reduces both CPU consumption (indicated by the numbers in the plot) and epoch duration (markers are closer to each other), 
   thus speeding up the overall learning process and reducing its cost.\\
 {\em Observation 2:} Larger values of~$K_1$ (solid lines) are associated with better performance after pruning.\\
{\em Observation 3:} Using lower-capability (``bronze'') nodes after pruning (warm colors) results in a larger difference between the learning performance  obtained when $K_1$~is small (i.e., 5) and when $K_1$~is larger (i.e., 25). Thus, achieving better performance while exploiting 
lower-capability nodes requires    switching model later.  

In a nutshell, switching from a model to another  may have significant benefits in terms of time and resource consumption;  however, its effects are hard to capture and foresee. 
Furthermore, the benefits of involving additional, yet heterogeneous, nodes depend upon the 
chosen models 
and the time at which to switch between  them. 
Thus, 
it is necessary to make all the decisions on model/nodes switching jointly, accounting for their interactions through a comprehensive  system model.

\section{System Model and Problem Formulation\label{sec:model}}
We now build the representation of the system we tackle,  
and formulate the problem of optimally matching DNN compression and training with resources/data availability. 

\subsection{Model components}
We envision a networked system for the compression and training of ML models where different nodes or sets of nodes are available, each characterized by  computational  and energy resources, and local datasets. A {\em learning orchestrator} controls the learning process. The system has two main components:
\begin{itemize}
    \item DNN {\em models}~$m\in\Mc$  
    that can be used for the training process; each  model is obtained by compressing the original DNN with a given technique or pruning ratio;
    \item sets~$n\in\Nc$ of {\em nodes} that can participate in the cooperative learning process.  
\end{itemize}

Let $k$ indicate the current epoch, $\ell(k)$ the loss function at $k$, and $T(k)$ the time at which epoch~$k$ finishes. 
$\Delta T(k)$ and $\Delta\ell(k)$
represent, respectively, the time taken by epoch~$k$, and the variation  in the value of  loss
 function it yielded. 
Finally, $\Delta E(k)$~denotes the {\em energy} consumed to perform epoch~$k$, and $E(k)$~the cumulative energy consumption until~$k$. 

Importantly, time and energy variations depend on the model ($m(k)$) and the set of nodes ($n(k)$) used at epoch $k$; 
also, they include {\em two} components each, i.e.,
\begin{align}
\Delta T(k)&{=}\tau^\text{change}(m(k{-}1),n(k{-}1),m(k),n(k))\nonumber\\&+\tau^\text{run}(m(k),n(k)),\label{eq:delta-t}\\
\Delta E(k) &{=}\epsilon^\text{change}(m(k{-}1),n(k{-}1),m(k),n(k))\nonumber\\&+ \epsilon^\text{run}(m(k),n(k))\label{eq:delta-e}.
\end{align} 
In the equations above, $\tau^\text{run}$ ($\epsilon^\text{run}$) represents the time (energy) to execute a given model over a set of nodes (hence, with the associated datasets), while  $\tau^\text{change}$  ($\epsilon^\text{change}$)~represents the time (energy) to change (i.e., {\em switch}) the model or nodes. In fact, model change  implies compressing the model, which may take time and energy, while a change in the set of nodes contributing to learning  requires  transferring the model.
Furthermore, not all model/nodes choices are possible, 
which is reflected by setting 
$\tau^\text{change}$, $\epsilon^\text{change}$, $\tau^\text{run}$, and $\epsilon^\text{run}$  to~$\infty$. 

The {\em evolution} of the loss function is given by:
\begin{multline}
\label{eq:delta-ell}
\Delta\ell(k){=}\lambda^\text{change}(k,m(k{-}1),m(k))\\+\lambda^\text{run}(k,m(k),n(k)).
\end{multline} 
Again, \Eq{delta-ell} includes two components: $\lambda^\text{change}$ -- the contribution\footnote{This is measured through a forward pass after model pruning.} of transitioning from the previous to the current model (if a model switch is performed), and $\lambda^\text{run}$ -- the effect of training that model for an epoch.
The sum of these components gives the difference between the loss at the current epoch~$k$ and that of  epoch~$k-1$, i.e., the result of the action we enact at epoch~$k$. 
However, now 
the two components may have different signs:  $\lambda^\text{run}\leq 0$ (the loss decreases) in most cases, while it is possible that~$\lambda^\text{change}\geq 0$, as changing model may increase the loss value \cite{yvinec2021red,gong2020privacy}. Impossible transitions between learning settings are associated with $\lambda^\text{change}=\infty$. 

In the following, when no confusion arises, we will drop the dependency of decision variables~$m$ and~$n$ from the epoch.
%

\subsection{Problem definition\label{sub:in-output_obj}}
 Given the impelling need to make ML sustainable \cite{energy,zaw2021energy}, our goal is to minimize the  overall learning energy consumption, while ensuring that the loss drops below a target value~$\ell^{\max}$ within  time~$T^{\max}$.
Specifically, for each epoch~$k$, the learning orchestrator has to select (i) which model~$m(k)$ to train in epoch $k$, and (ii) which set~$n(k)$ of nodes to involve next in the learning process. 
Based on these decisions, the values~$\Delta T(k)$, $\Delta E(k)$, and~$\Delta \ell(k)$~follow, 
expressing, respectively, how long  iteration $k$ takes, how much energy it consumes, and what improvement in the learning it yields. 

The learning orchestrator acts based on the knowledge of the characteristics of the network nodes that can contribute to a learning process, and of the computational, temporal, and energy impact of running a model. Such values can indeed be calculated following, e.g., the methodology in~\cite{abtahi2017accelerating}. Thus, sets~$\Mc$ and~$\Nc$, as well as functions 
$\tau^\text{run}$, $\tau^\text{change}$, $\epsilon^\text{run}$ and $\epsilon^\text{change}$, are given from the viewpoint of our problem.   

On the contrary,  $\lambda^\text{run}$ and~$\lambda^\text{change}$ can only be estimated by the learning orchestrator, through estimators $\hat{\lambda}^\text{run}$ and~$\hat{\lambda}^\text{change}$.  This reflects the fact that understanding how training a specific model over specific nodes (hence, also data) improves learning is a hard problem, 
and, indeed, all existing works merely provide approximations and/or bounds to such quantities. In the following, we treat those estimators as given; then,  in \Sec{estimating} we present the methodology used at the learning orchestrator in order to compute them. 

Owing to the discrete-time, combinatorial nature of the problem, 
we propose an {\em approximate dynamic programming} (ADP) 
formulation thereof, as described below.
Dynamic programming is indeed well-suited to cope with combinatorial problems where the system state evolves over time and the same decision process shall be repeated for multiple epochs.

\subsection{ADP formulation}

First, we define  the state space, set of actions, and cost function. The state at epoch~$k$ is given by~$\sb(k)=\left(k,\ell(k),T(k),m,n\right)$, 
while 
the set of actions available from  state~$\sb(k)$ is given by all possible decisions~$(m',n')\in\Mc\times\Nc$ 
such that   
the switch they entail (if any) is feasible.     
The {\em cost function}~$\Cb(\sb(k),\ab(k))$  expresses the (immediate) cost of executing action~$\ab$ while in state~$\sb$ at epoch~$k$, as the corresponding consumed energy  $\Cb(\sb(k),\ab(k))=\Delta E(k)$. 
Such a cost  comes directly from  \Eq{delta-e}, i.e., $\Cb(\sb(k),\ab(k))=\epsilon^\text{change}(m,n,m',n')+\epsilon^\text{run}(m',n')$.

The {\em value function}~$\Vb(\sb(k))$, i.e., how desirable it is to be in state~$\sb(k)$, requires a more sophisticated, and domain-specific, definition.  
We set the value of being in  state~$\sb(k)$ equal to 0 when, after $T^{\max}$, the loss is above $\ell^{\max}$; we set it to the maximum value (i.e., 1) whenever $\ell(k)< \ell^{\max}$ while $T(k)\leq T^{\max}$. 
For all other states, we compare the current loss~$\ell(k)$ and 
time~$T(k)$ with an {\em ideal} loss-time curve $\ell^\text{ideal}(t)$ which:
(i) starts at~$\ell(0)$ for~$T=0$;
(ii) ends at~$\ell^{\max}$ for~$T=T^{\max}$, and
    (iii) follows a power law in the between.
The latter comes from the finding invariably reported in both theoretical~\cite{oymak2020toward,saxe2013exact,allen2019learning} and 
experimental~\cite{hestness2017deep} works. Then, we can write
the value of being in state~$\sb(k)$ as  the difference between ideal and real loss values, i.e., 
\begin{equation}
\label{eq:v-log}
\Vb(\sb(k))=\mathsf{logistic}\left(\ell^\text{ideal}(T(k))-\ell(k)\right),
\end{equation}
where the value is normalized 
via a logistic function.

Dynamic  programming problems can  be solved {\em in principle} by optimizing Bellman's equation, 
i.e., choosing at each epoch the action 
minimizing the total energy cost:
\begin{eqnarray}
\label{eq:bellman}
&& \min_{\ab(k)\in\Ab^k} \sum_k \Cb(\sb(k),\ab(k)) \\
&& \mbox{s.t.} \,\,\,\Vb(\sb(K))=1 \,\,\,;\,\,\, T(K)\leq T^{\max} \,.
\end{eqnarray} 
To solve our problem in real-world scenarios, however, there are two major challenges to face. First, the learning orchestrator does not have access to the future decrease (or increase) in the loss value~$\Delta\ell(k)$, and how our decisions influence it. 
A possible solution to this issue is to use traditional Deep Reinforcement Learning (DRL) approaches. For instance, Deep Q-Learning algorithms would implicitly learn the probabilistic dynamics of loss as a function of taken actions. However, training DRL agents often requires very large datasets to achieve satisfactory convergence, and may result in weak generalization. Herein, we take a different approach, where we build an ADP framework based on low-complexity neural networks (NN) estimators of possible loss trajectories with a finite time horizon. 
Second, in view of the number of possible actions, the learning orchestrator has to identify a subset of actions to evaluate at each epoch. Such challenges are dealt with in \Sec{estimating} and \Sec{buddy}, respectively.

\section{Estimating the Performance of Learning}
\label{sec:estimating}

As discussed in Sec.\,\ref{sub:in-output_obj}, neither of the quantities contributing to the loss evolution ($\lambda^\text{change}(k,m,m')$ and~$\lambda^\text{run}(k,m,n)$) is known exactly.  We thus introduce  {\em estimators} for $\Delta \ell(k)$. Specifically, for~$\lambda^\text{run}(k,m,n)$: 
\begin{itemize}
    \item an {\em expected-value} estimator~$\hat{\lambda}^\text{run}_\text{exp}(k,m,n)$ of the loss reduction value;
    \item a {\em robust} estimator~$\hat{\lambda}^\text{run}_\text{rob}(k,m,n)$, such that  $\lambda^\text{run}(k,m,n) \leq\hat{\lambda}^\text{run}_\text{rob}(k,m,n)$ with high probability. 
\end{itemize}
In general, $\hat{\lambda}^\text{run}_\text{exp}(k,m,n)\leq\hat{\lambda}^\text{run}_\text{rob}(k,m,n)$, i.e., 
the robust estimator is the most pessimistic. 
Likewise, for~$\lambda^\text{change}(k,m,m')$,  we can introduce the corresponding  estimators,~$\hat{\lambda}^\text{change}_\text{exp}(k,m,m')$ and $\hat{\lambda}^\text{change}_\text{rob}(k,m,m')$,  with similar properties.

\begin{figure}[t]
\center
\includegraphics[width=\columnwidth]{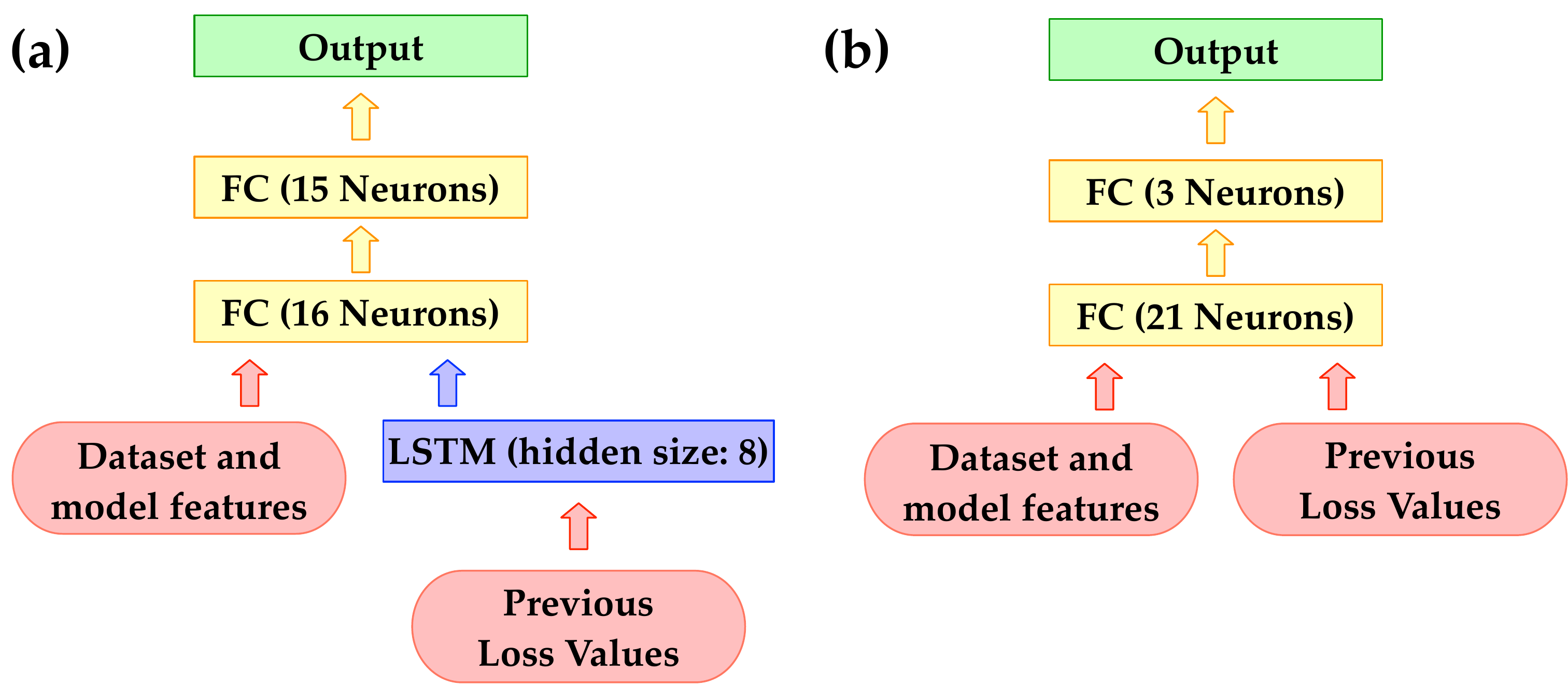}\vspace{-2mm} 
\caption{Architecture of the $\hat{\lambda}^\text{run}$ (a) and $\hat{\lambda}^\text{change}$ (b) estimators.}
\label{fig:nn} \vspace{-4mm}
\end{figure}

To obtain both the expected-value and the robust estimator, the learning orchestrator leverages the knowledge of the number of classes of the datasets owned by the  nodes and makes use of NN architectures that can predict the expected training loss variation as well as determine the prediction uncertainty. Specifically, for  $\lambda^\text{run}(k,m,n)$, we take as a starting point the Long Short-Term Memory (LSTM) model in \cite{altche2017lstm} and develop a similar, yet simpler, 
branched architecture, as depicted in Fig.~\ref{fig:nn}(a). 
The features fed to the first Fully Connected (FC) layer are the time-independent parameters, i.e., the number of classes and samples in the dataset of the  nodes set currently training the DNN model, and the pruning ratio $F$ of the current model. 
The input of the LSTM layer is the sequence of loss values obtained so far in the DNN model. 
The NN predicts  the expected  value of $\lambda^\text{run}(k,m,n)$  as well as two associated quantiles (namely,  $0.05$ and $0.95$), yielded by the learning process in the next 5 epochs (thus, the FC layer output size is 15, i.e.,  number of predicted metrics times  number of prediction steps). So doing, we obtain  $\hat{\lambda}^\text{run}_\text{exp}(k,m,n)$ and 
$\hat{\lambda}^\text{run}_\text{rob}(k,m,n)$, with the latter given by the 0.95 quantile.

As for $\lambda^\text{change}(k,m,m')$, since the goal is to predict the loss variation when we move from one DNN model to another, we leverage regression, using the NN in Fig.~\ref{fig:nn}(b). 
The NN is fed the pruning ratio and the $5$ loss values preceding the model switch.  
The regression model predicts  the expected value $\hat{\lambda}^\text{change}_\text{exp}(k,m,m')$ as well as the 0.05 and 0.95 quantiles in the next epoch of the DNN training, with  $\hat{\lambda}^\text{change}_\text{rob}(k,m,m')$ being again the 0.95  quantile.

\begin{figure}[t]
\center
\includegraphics[width=0.8\columnwidth]{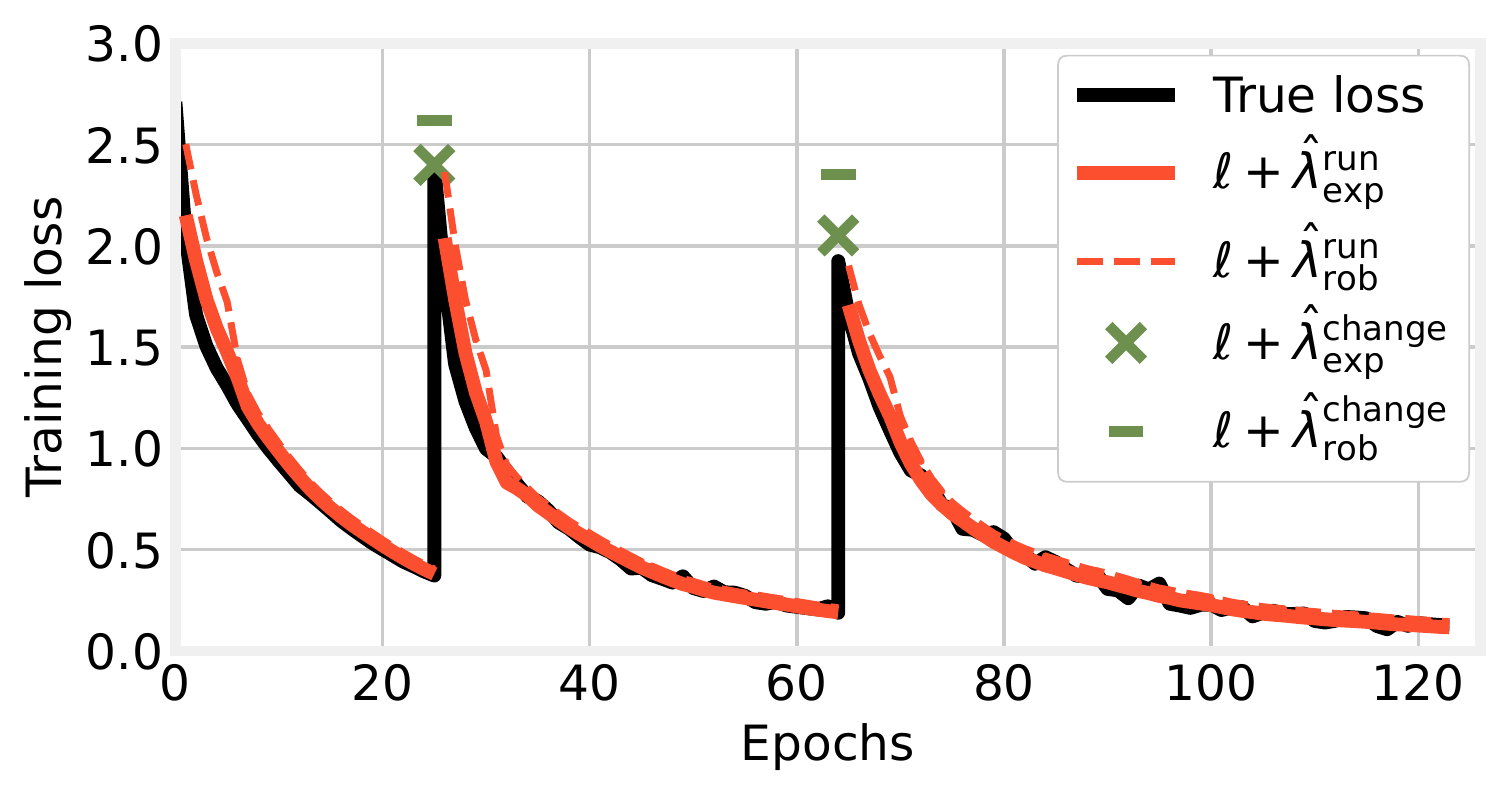} 
\caption{Example of true loss  vs expected-value and robust estimators.}
\label{fig:loss} \vspace{-4mm}
\end{figure}

As depicted in 
\Fig{loss} the above estimators produce very accurate predictions (red lines and green markers for $\lambda^\text{run}$ and $\lambda^\text{change}$, resp.) of the true loss (black line). The figure refers to  the training loss of the VGG11 DNN model initially trained with 45,000 samples of the CIFAR-10 dataset with 10 classes. After  $25$ epochs, the model is pruned with $F_1=0.75$ and handed over to a set of nodes owning  5,000 samples belonging to 13 classes, taken from the CIFAR-10 and CIFAR-100 datasets. Likewise, after  $40$ more epochs, the model is  further pruned with  $F_2 = 0.5$, and passed to a third set of nodes owing  1,500 samples belonging to 15 classes, taken from the same combination of datasets. 

Finally,  to improve the reliability of the robust estimator, the learning orchestrator compares the values obtained through the above NN to the lower bounds that are available  for~$\lambda^\text{run}(k,m,n)$~\cite[Theorem~1]{li2019convergence}  and  for~$\lambda^\text{change}(k,m,m')~$\cite[Sec.~3]{yvinec2021red}. If they result to be lower than the bounds, the latter are taken as robust estimators.

\section{The PACT Algorithm}
\label{sec:buddy}

The goal of PACT is to let the learning orchestrator efficiently find high-quality solutions to the problem in \Eq{bellman}, which, as shown later, is NP-hard. PACT consists of  three steps:
    
    1) Create an {\em expanded graph} representing the possible decisions and their outcome;
    
2)    Using such a graph, identify a set of decisions deemed {\em feasible} based on the estimated loss trajectory;
    
    3) By combining learning- and energy-related information,  choose the best feasible solution to enact.


{\bf Step 1: Expanded graph.} 
The expanded graph is a directed graph built according to the following rules:
\begin{itemize}
    \item The {\em vertices} represent the states of the system; they are labeled with the current epoch~$k$, model~$m(k)$ and set of nodes~$n(k)$ being used, and the total elapsed time~$T(k)$ and current loss~$\ell(k)$. 
    With the aim of identifying feasible solutions, the latter quantity is computed using the robust estimators $\hat{\lambda}^\text{change}_\text{rob}(k,m,m')$ and $\hat{\lambda}^\text{run}_\text{rob}(k,m,n)$;

    \item Elapsed time and loss values are represented, respectively, with resolutions~$\gamma_T$ and~$\gamma_\ell$ (e.g., if~$\gamma_\ell=0.1$, a vertex with~$\ell=0.1$ or~$0.2$ can exist, but not with~$\ell=0.15$);

    \item A directed {\em edge} is drawn between two vertices if there is an action making  the system move from one  corresponding state to the other; each edge is labeled with the {\em energy consumption} of the associated action,  as     
    in \Eq{delta-e};
    
    \item Each vertex representing a feasible state of the system  (i.e., with $\ell(k) \leq \ell^{\max}$ and $T(k) \leq T^{\max}$) is further connected to a virtual node~$\Omega$ through a zero-cost edge.
\end{itemize}

The graph is created through the \textsc{CreateExpandedGraph} function, presented in \Alg{makegraph}. First, all {\em vertices} are created, representing all valid combinations of model and set of nodes,  epoch, loss value, and elapsed time (\Line{mkg-forall-mn}--\Line{mkg-forall-t}).
Note that 
the quantization parameters $\gamma_\ell$ and~$\gamma_T$   (\Line{mkg-forall-k}--\Line{mkg-forall-t}) allow us to control the trade-off between size of the graph and quantization error.

For each vertex~$v$, 
the effect of taking action~$\ab$ from vertex~$v$ is determined by computing the resulting elapsed time  and the required energy (\Line{mkg-compute-t1}--\Line{mkg-compute-e}).
If either is infinite,  
then  taking action~$\ab$ while in state~$v$ is impossible, and we move on to the next action. 
Otherwise, the loss~$\ell'$ resulting from taking the action is computed using the robust estimator (\Line{mkg-compute-l1}).
Now, tuple~$(k+1,m',n',\ell',T)$ would describe the state the system lands on after performing~$\ab$ from~$v$; however, due to the way  the vertices are created  (i.e., using $\gamma_\ell$ and~$\gamma_T$),  
such a tuple may not correspond to a vertex in~$\Vc$. Accordingly, in \Line{mkg-fix-l1}--\Line{mkg-fix-t1},
$\ell'$ and~$T'$ are cast into 
integer multiples of~$\gamma_\ell$ and~$\gamma_T$. Then,  vertex~$v'$ representing the new state is identified (\Line{mkg-create-v1}), 
and an edge from~$v$ to~$v'$ is added 
using the appropriate energy value~$E$ 
as its weight. Finally, if~$v$ is feasible,  
$v$ is connected  to~$\Omega$
(\Line{mkg-connect-omega}).

\begin{algorithm}[t]
\caption{
Creating the expanded graph\label{alg:makegraph}
} 
\begin{algorithmic}[1]
\Function{CreateExpandedGraph}{}

\State{$\Vc\gets\{\Omega\}$} \Comment{set of vertices} \label{line:mkg-omega}
\ForAll{$m\in\Mc,n\in\Nc$} \label{line:mkg-forall-mn}
 \ForAll{$k\in\left[1,2,\dots,\ceil*{\frac{T^{\max}}{\gamma_T}}\right]$} \label{line:mkg-forall-k}
  \ForAll{$\ell\in[0,\gamma_\ell,2\gamma_\ell,\dots,\ell(0)]$} \label{line:mkg-forall-l}
   \ForAll{$T\in[0,\gamma_T,2\gamma_T,\dots,T^{\max}]$} \label{line:mkg-forall-t}
    \State{$v\gets(k,m,n,\ell,T)$} \label{line:mkg-create-v}
    \State{$\Vc\gets\Vc\cup\{v\}$} \label{line:mkg-add-v}
   \EndFor
  \EndFor
 \EndFor
\EndFor

\State{$\Ec\gets\emptyset$} \Comment{set of edges} 
\ForAll{$v=(k,m,n,\ell,T)\in\Vc$} \label{line:mkg-forall-v}
 \ForAll{$\ab=(m',n')\in\Ab$} \label{line:mkg-forall-a}
  \State{$T'\gets T+\tau^\text{change}(m,n,m',n')+\tau^\text{run}(m,n)$} \label{line:mkg-compute-t1}
  \State{$E\gets\epsilon^\text{change}(m,n,m',n')+\epsilon^\text{run}(m',n')$} \label{line:mkg-compute-e}
  \If{$T'>T^{\max} \vee E=\infty$} \label{line:mkg-check-infty}
   \State{$\textbf{continue}$} \Comment{infeasible, skip this action} \label{line:mkg-continue}
  \EndIf
  \State{$\ell'\gets\ell+\hat{\lambda}^\text{change}_\text{rob}(k,m,m')+\hat{\lambda}^\text{run}_\text{rob}(k,m,n')$} \label{line:mkg-compute-l1}
  \State{$\ell'\gets\gamma_\ell\ceil*{\frac{\ell'}{\gamma_\ell}}$} \label{line:mkg-fix-l1}
  \State{$T'\gets\gamma_T\ceil*{\frac{T'}{\gamma_T}}$} \label{line:mkg-fix-t1}
  \State{$v'\gets (k+1,m',n',\ell',T')$} \label{line:mkg-create-v1}
  \State{$\Ec\gets\Ec\cup\{(v,v',\textsf{weight}=E)\}$} \label{line:mkg-addedge}
 \EndFor
 \If{$\ell\leq\ell^{\max}\wedge T\leq T^{\max}$} \label{line:mkg-check-feasible}
  \State{$\Ec\gets\Ec\cup\{v,\Omega\}$} \Comment{feasible state} \label{line:mkg-connect-omega}
 \EndIf
\EndFor

\State\Return{$\Gc=(\Vc,\Ec)$} \label{line:mkg-return}
\EndFunction
\end{algorithmic}
\end{algorithm}

\begin{figure}[b]
\centering
\includegraphics[width=0.95\columnwidth]{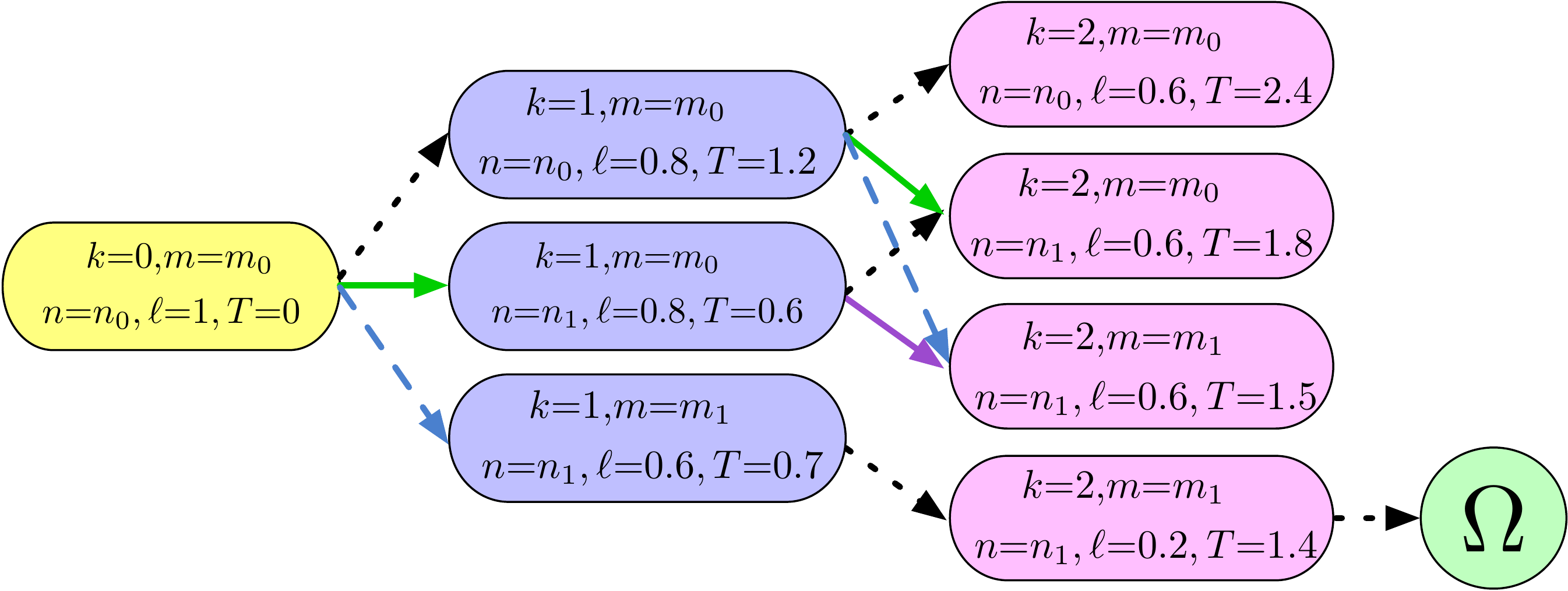}
\caption{Example of the expanded graph generated by PACT, with resolution values $\gamma_T=0.1$ and~$\gamma_\ell=0.1$, learning target $\ell^{\max}=0.25$, and time limit $T^{\max}=1.5$. Edge colors correspond to switches made across subsequent epochs: node only (solid green), model only (solid purple), both (dashed blue), neither (dotted black).
    \label{fig:expanded}\vspace{-4mm}
}
\end{figure}

\Fig{expanded} presents an example of expanded graph. The initial vertex is associated with epoch~$k=0$, model~$m(0)=m_0$, node~$n(0)=n_0$, loss~$\ell(0)=1$ and elapsed time~$T(0)=0$. The learning target is~$\ell^{\max}=0.25$ and the time limit is~$T^{\max}=1.5$. Also, the resolution values are set to~$\gamma_T=0.1$ and~$\gamma_\ell=0.1$. From the current state, it is possible to change the node (switching to more capable~$n_1$), model (switching quicker-converging $m_1$), both, or neither; such actions are represented (resp.) by solid green, solid purple, dashed blue, and dotted black edges in the figure. Different combinations of possible switches yield different combinations of loss and elapsed time, only one of which -- the bottom, pink  vertex -- is feasible, hence, connected to~$\Omega$.

{\bf Step 2: Feasible paths.} 
Next, PACT uses the expanded graph to identify a set of  paths deemed {\em feasible}; the first edge of such paths represents a feasible action. 
To mitigate the impact of potential errors in the loss estimation (which in principle may jeopardize feasibility),  
the expanded graph is built using the {\em robust} estimators of the loss variation, which 
guarantees that all paths landing at a feasible node are, indeed, feasible with high probability. 
Thus, using function \textsc{FindFeasiblePaths} in \Alg{feas}, PACT seeks for paths that (i) start from the current state, and (ii) arrive to a feasible state, i.e., to a vertex connected to~$\Omega$. 
Specifically, for each vertex~$v$ corresponding to a feasible state,  
it determines the shortest path (\Line{feas-shortest}) 
from the current state~$v_\text{curr}$ 
to~$v$. Such paths are collected in set~$\Pc$ and associated with a weight corresponding to the sum of weights (i.e., energy consumption) of their edges. 

\begin{algorithm}[t]
\caption{
Finding feasible paths\label{alg:feas}
} 
\begin{algorithmic}[1]
\Function{FindFeasiblePaths}{}
\State{$v_\text{curr}\gets(k,m,n,\ell,T)$} \label{line:feas-v-curr}
\State{$\Pc\gets\emptyset$} \Comment{feasible paths} \label{line:feas-create-pc}
\ForAll{$v$: $(v,\Omega)\in\Ec$} \label{line:feas-forall-omega}
 \State{$p\gets\mathbf{shortestPath}(v_\text{curr},v)$} \label{line:feas-shortest}
 \State{$\Pc\gets\Pc\cup\{p,\text{weight}=\sum_{e\in p}\text{weight}[e]\}$}  \label{line:feas-add}
\EndFor
\State\Return{$\Pc$}
\EndFunction
\end{algorithmic}
\end{algorithm}

{\bf Step 3: Making the best decision.} 
Once the set of feasible paths, and associated feasible actions, has been identified, using robust estimators to choose the decision to enact would be overly cautious, possibly resulting in unnecessarily higher energy costs. Thus, 
PACT accounts for two additional aspects when selecting an action: an {\em opportunity} and a {\em risk} factor.  Such factors and the  path weight  are integrated into a {\em score}, and the action corresponding to the lowest score is enacted.

For every 
path~$p\in\Pc$, scores are computed in the \textsc{ChooseAction} function  in \Alg{choose}. The opportunity factor, $\texttt{opp}\geq 1$, is given by the ratio of  (i) the sum of the expected loss to (ii) the sum of the robust loss associated with the edges in~$p$ (\Line{choose-compute-opp}). 
The intuition is to make it easier to choose actions with a good expected loss, since the robust estimator may be too pessimistic. 
As for the risk factor, its high-level purpose is to avoid undoing decisions. To this end, PACT seeks for paths on the expanded graph that lead from the first node of~$p$, to a vertex~$\overline{v}\in\overline{\Vc}$ associated with the current model~$m$ 
(\Line{choose-vbar}), 
and thence to~$\Omega$. 
The risk factor, $\texttt{risk}\geq 1$, associated with path~$p$ is then computed in \Line{choose-risk} 
as the ratio
of the minimum among the weights of such paths
to the weight of~$p$ (defined in \Alg{feas}).

The score of path~$p$ is obtained 
in \Line{choose-combine} 
as $p$'s weight, divided by the opportunity factor, and multiplied by the risk factor. Then the action associated with the minimum-score path  is returned.  
It is important to underline that the shortest path going from the current state to~$\Omega$ represents the  lowest-cost decision since edge weights are set to the energy cost of the corresponding actions. Thus, the ultimate outcome of this step is the action with the lowest energy cost to enact.
\begin{algorithm}[t]
\caption{
Choosing the next action\label{alg:choose}
} 
\begin{algorithmic}[1]
\Function{ChooseAction}{}
\State{$\texttt{scores}\gets\{\}$}

\ForAll{$p\in\Pc$}
 \State{$\overline{w}\gets 0$} \Comment{opportunity} \label{line:choose-wbar}
 \State{$\texttt{Le}\gets 0\quad ;\quad \texttt{Lr}\gets 0$} \label{line:choose-init}
 \ForAll{$((k,m,n,\ell,T),(k',m',n',\ell',T'))\in p$} \label{line:choose-forall-p}
  \State{$\texttt{Le}{\gets}\texttt{Le}{+}\hat{\lambda}^\text{change}_\text{exp}(k,m,m'){+}\hat{\lambda}^\text{run}_\text{exp}(k,m,n')$} \label{line:choose-update-lexp}
  \State{$\texttt{Lr}{\gets}\texttt{Lr}{+}\hat{\lambda}^\text{change}_\text{rob}(k,m,m'){+}\hat{\lambda}^\text{run}_\text{rob}(k,m,n')$} \label{line:choose-update-lrob}
 \EndFor
 \State{$\texttt{opp}\gets\texttt{Le}/\texttt{Lr}$} \label{line:choose-compute-opp}

 \State{$\overline{V}\gets\{v\in\Vc\colon v[1]=m\}$} \Comment{risk} \label{line:choose-vbar}
 \State{$\texttt{Wr}{\gets}\min_{\overline{v}{\in}\overline{\Vc}}\text{weight}(\textbf{shortestPath}(p[1]{,}\Omega{,}\text{via }\overline{v})$} \label{line:choose-wr}
 \State{$\texttt{risk}\gets\texttt{Wr}/\text{weight}[p]$} \label{line:choose-risk}
 \State{$\texttt{scores}[p]\gets\text{weight}[p]\cdot\texttt{risk}/\texttt{opp}$} \label{line:choose-combine}
\EndFor

\State{$p^\star\gets\arg\min_{p\in\Pc}\texttt{score}[p]$} \label{line:choose-bestpath}
\State\Return{$\ab=(p[1][1],p[1][2])$} \label{line:choose-return}

\EndFunction
\end{algorithmic}
\end{algorithm}

%

\subsection{Problem and algorithm analysis}
\label{sec:analysis}

\begin{property}
The problem of optimizing \Eq{bellman} is NP-hard.
\label{prop:hard}
\end{property}
The proof is based on a reduction in polynomial time from  the generalized assignment problem (GAP)~\cite{cattrysse1992survey}, which is known to be NP-hard. 
Furthermore, we prove that: 
\begin{property}
PACT's time complexity is polynomial.
\label{prop:polynomial}
\end{property}
\begin{IEEEproof}
PACT's complexity is given by the sum of the complexity of \Alg{makegraph}--\Alg{choose}. In \Alg{makegraph}, the first loop is run at most~$|\Vc|=MN\ceil*{\frac{T^{\max}}{\gamma_T}}^2\ceil*{\frac{\ell(0)}{\gamma_\ell}}$ times, and the second one for at most~$|\Vc|MN$ times. 
\Alg{feas} computes at most~$|\Vc|^2$ shortest paths, each of which (e.g., using Dijkstra's algorithm~\cite{fredman1976new}) incurs a polynomial complexity.  \Alg{choose} iterates over set~$\Pc$ of feasible paths, whose number cannot exceed $|\Vc|$ (as per \Alg{feas}, \Line{feas-forall-omega}).  
Thus, \Alg{makegraph} represents the dominating contribution to PACT's  complexity, which proves the thesis.
\end{IEEEproof}
Importantly, \Prop{polynomial} concerns the {\em worst-case} time complexity of PACT, which in  practice has  substantially lower complexity. In particular, the shortest-path routines  used in \Alg{feas} and \Alg{choose} have been heavily optimized, and perform very efficiently in practice~\cite{fredman1976new}.

At last, we prove the following property about 
 how good PACT's solutions are at minimizing the objective in \Eq{bellman}. 
\begin{property}
If predictions are exact, their time horizon is sufficiently long, and
all~$\Delta\ell$ and~$\Delta T$ values are integer multipliers of~$\gamma_\ell$ and~$\gamma_T$, then PACT is optimal.
\label{prop:opti}
\end{property}
\begin{IEEEproof}
The proof comes from inspection of \Alg{makegraph}--\Alg{choose}, which consider all possible options, 
hence, no feasible solutions are 
ignored. Further, the shortest-path problem  in \Alg{feas} and \Alg{choose} can be efficiently solved to the optimum. 
If the hypothesis holds, then the ceiling operators in \Alg{makegraph} (\Line{mkg-fix-l1} and \Line{mkg-fix-t1}) have no effect, hence, there is no possible source of suboptimality.
\end{IEEEproof}
An important consequence of \Prop{opti} is that, by varying~$\gamma_\ell$ and~$\gamma_T$, we can effectively trade off how close  to the optimum the solution gets with PACT's time  complexity.

\section{Performance Results}
\label{sec:results}

We first describe in \Sec{loss-prediction} how we implement the loss prediction. Then we compare the performance of PACT against the optimum and  state-of-the-art approaches in \Sec{res}, under the scenario and settings described in \Sec{ref}. 

\subsection{Loss prediction implementation\label{sec:loss-prediction}} 
To collect the training loss data necessary for the training of the estimators for $\lambda^\text{change}$ and $\lambda^\text{run}$, we consider a scenario with three sets of nodes: the gold set  has  45,000 samples of the CIFAR-10 belonging to 10 classes; the silver one has  5,000 samples out of the CIFAR-10 and CIFAR-100 datasets, belonging to 13 classes; and the bronze set has  1,500 samples out of the CIFAR-10 and CIFAR-100 datasets, belonging to 15 classes. 
Experiments always start with the gold set of nodes training  a full model.  Then both the
cases of one and two pruning occurrences (i.e., two and three models) are considered, with pruning being performed as described in \Sec{motivating}. 
In the former case, after $K_1$ epochs, the model is pruned with pruning ratio $F_1$ and hand it over to the silver or the bronze set, which continues the training. In the second case, after the second set of nodes trains the pruned model for $K_2$ more epochs, and then prunes the model again with ratio $F_2$, it sends it to the last set of nodes,  which completes the training.
Experiments have been run for all  combinations of pruning/training and of the  involved parameters; specifically, we have considered: $K_1, K_2 \in [5,15,25,40,60,100]$,  $F_1 \in [0.5,0.75]$ and $F_2 \in [0.25, 0.5]$.
Notice that these are the combinations we leverage for our training, and do not limit in any way the decisions that PACT or its benchmarks can make. 
For the training of the NNs used for prediction, 
we used the Adam optimizer with learning rate of $10^{-3}$, and set the batch size to 16. The loss is given by the Mean Square Error (MSE), to which  the so-called tilted loss term has been added, to compute the quantiles~\cite{rodrigues2020beyond}.  

To assess the quality of prediction, we evaluate: the Mean Absolute Error (MAE), 
the Mean Interval Length (MIL) (i.e., the average width of the prediction interval), and the  Interval Coverage Percentage (ICP) (i.e., the fraction of true values falling within the relative prediction interval), with the latter two  indicating the quality of the quantiles prediction.  
The excellent results we get are presented in \Tab{results}, which  reports the mean and the standard deviation  of the three metrics, computed over all possible combinations of pruning and training configurations, and executing 10 runs for each of such cases.  
MIL and ICP are calculated for the 90\% prediction interval, as the considered quantiles are 0.05 and 0.95. 

\subsection{Reference scenario}\label{sec:ref}

To assess PACT's 
performance, we focus on a smart factory-based application
using the 
VGG11 DNN. 
Three models are considered (called {\em L}, {\em M}, and {\em S}),  corresponding to (resp.) a full DNN, a DNN pruned with $F=0.5$, and a DNN further pruned with $F=0.75$.
Again, we consider that {\em gold},  {\em silver}, and {\em bronze} sets of nodes are available, located (resp.) in the cloud,   in the far edge of the network infrastructure, and   in the near edge covering the smart-factory premises. 
To match the model size with the nodes' capability, each  model best runs on one of the sets, hence, switching between models also implies changing the set of nodes to use. 
To  reflect the real-world capabilities of (resp.) NVIDIA Ampere A100~\cite{a100} (gold nodes), NVIDIA RTX A4000~\cite{rtxa4000} (silver nodes), and Raspberry Pi's Videocore~6~\cite{videocore6} (bronze nodes) GPUs,  (i)  the duration and energy cost required for $M$  to be trained by the silver set for one epoch are one fifth and a half (resp.) smaller than those experienced when the $L$ model is trained by the gold set, and (ii) such values reduce to a half and to one fifth (resp.)  when $S$ is trained by the bronze nodes. 
Further, for  simplicity, we set a very long time limit of~$T^{\max}=1,000$~time units.

\begin{figure*}
\centering
\includegraphics[width=.32\textwidth]{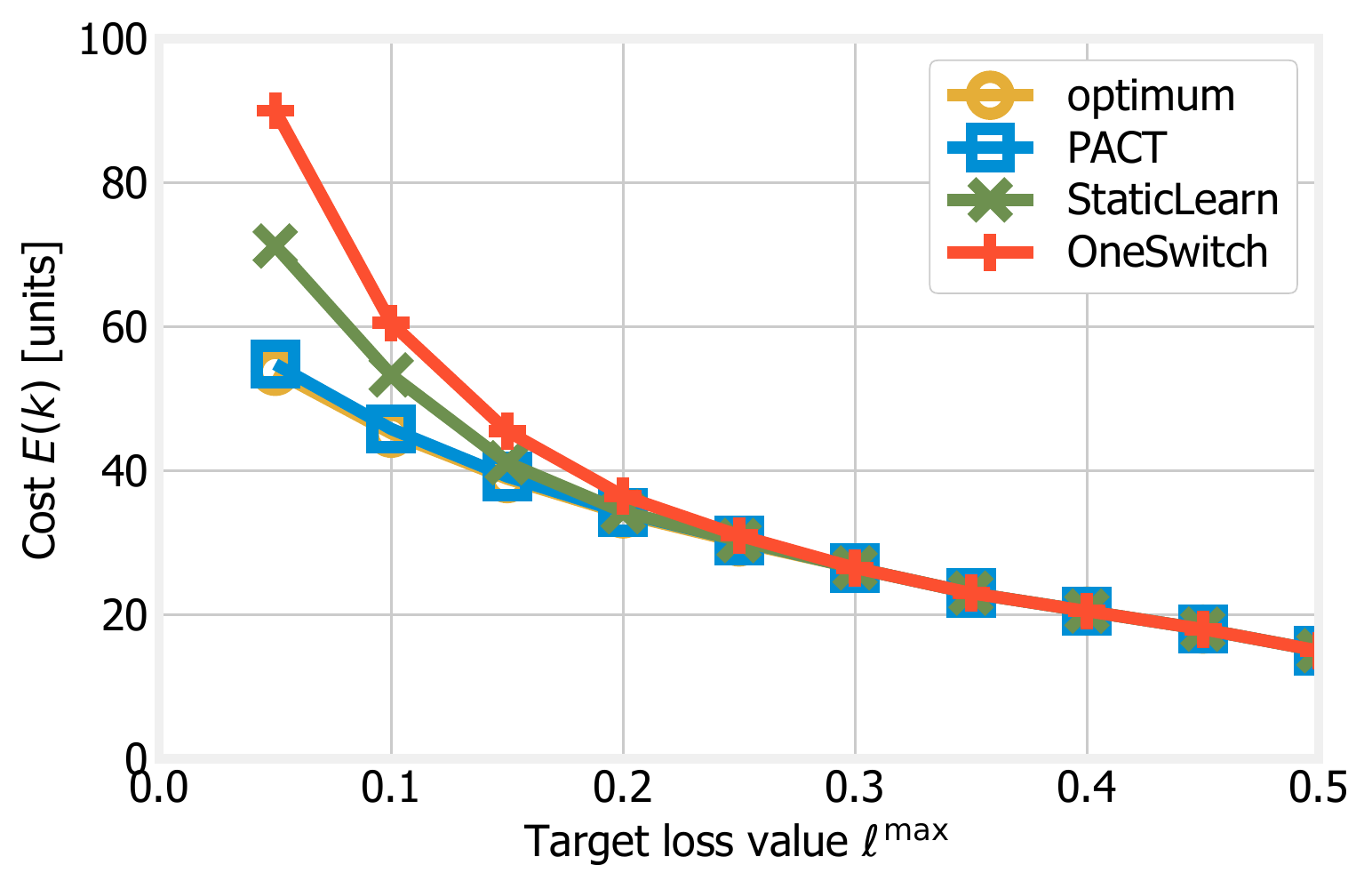}
\includegraphics[width=.32\textwidth]{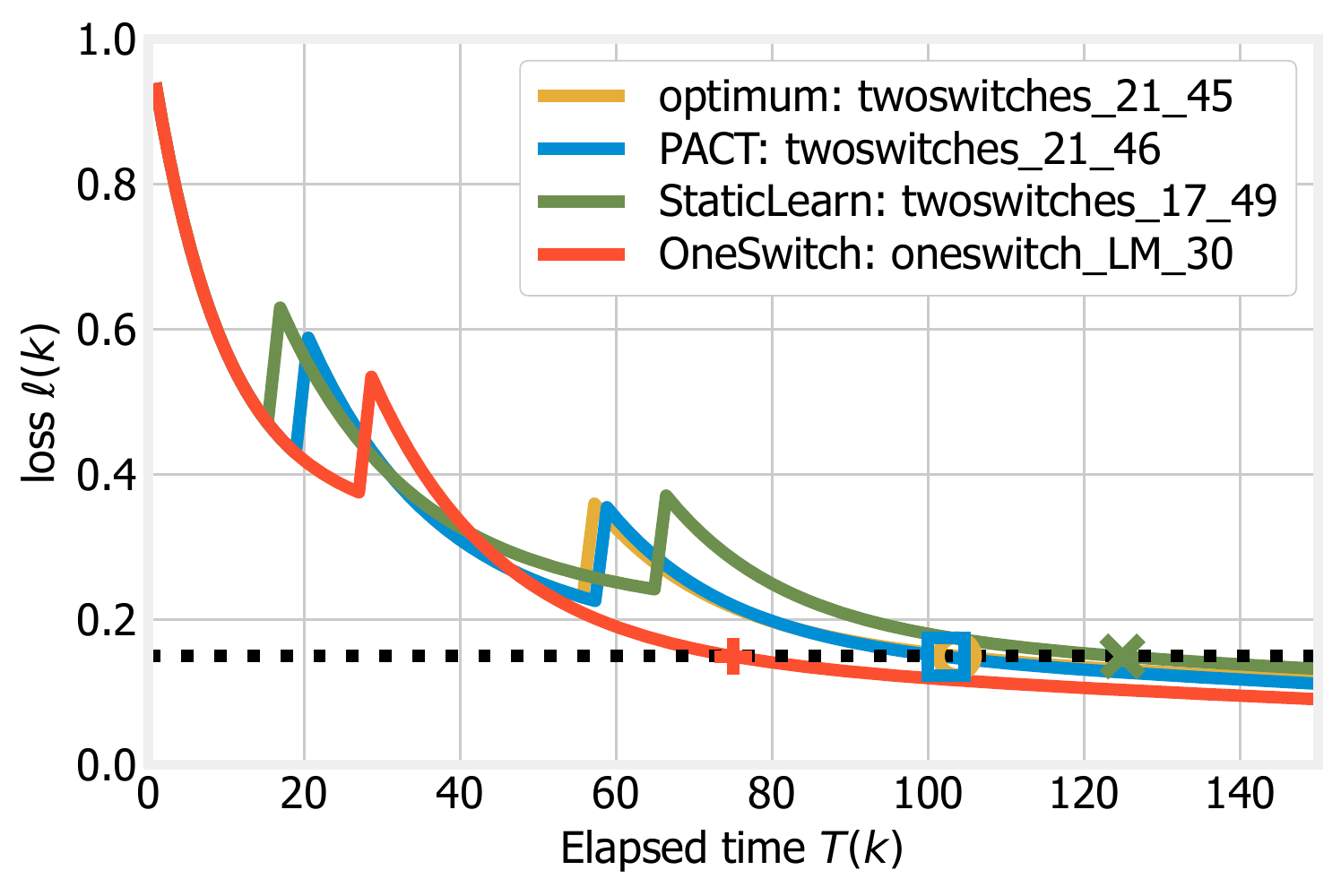}
\includegraphics[width=.32\textwidth]{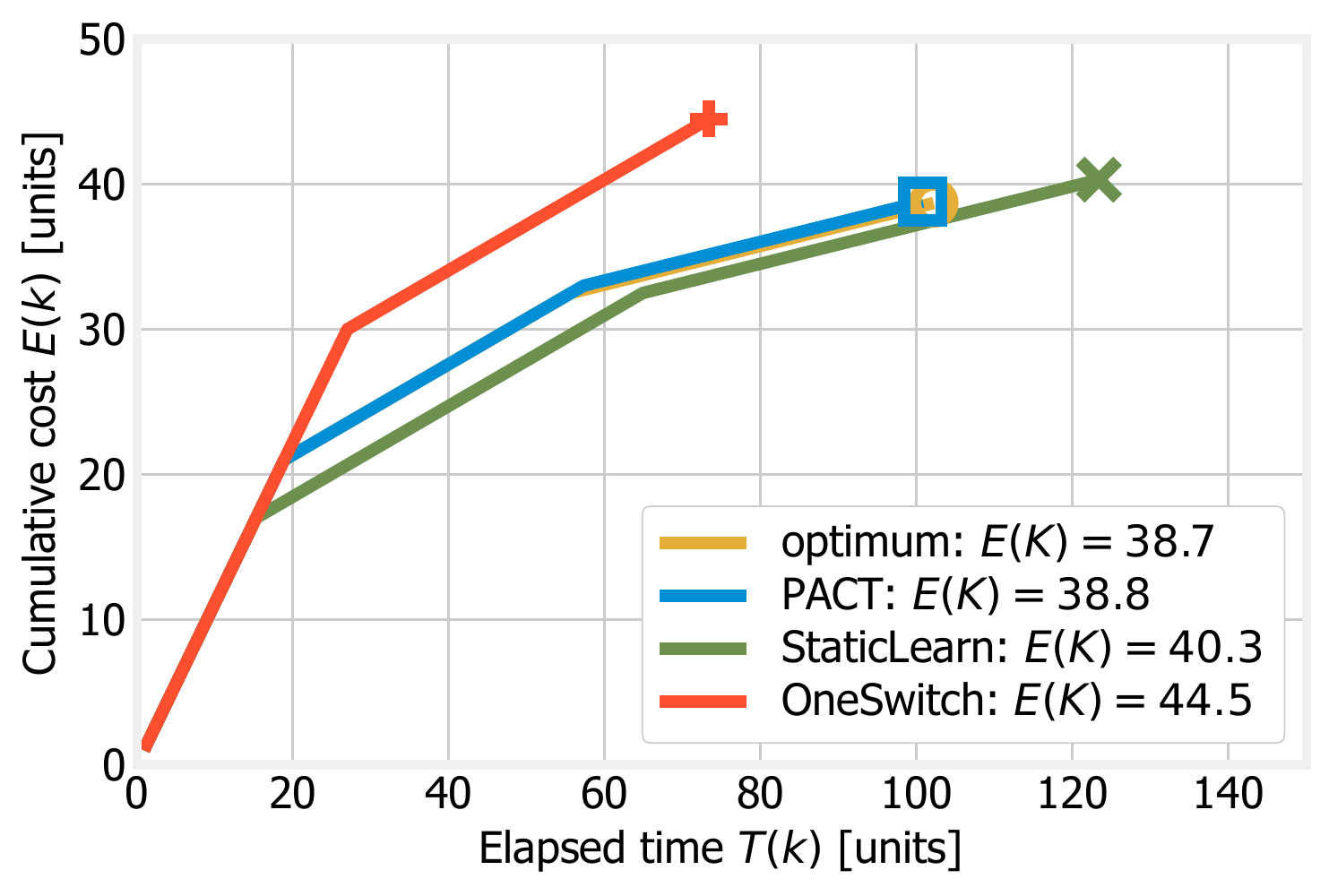}
\caption{
PACT and benchmark strategies: cost for different values of~$\ell^{\max}$ (left); evolution of the loss (center) and cost (right) when~$\ell^{\max}=0.15$. 
    \label{fig:strats}
} 
\centering
\includegraphics[width=.32\textwidth]{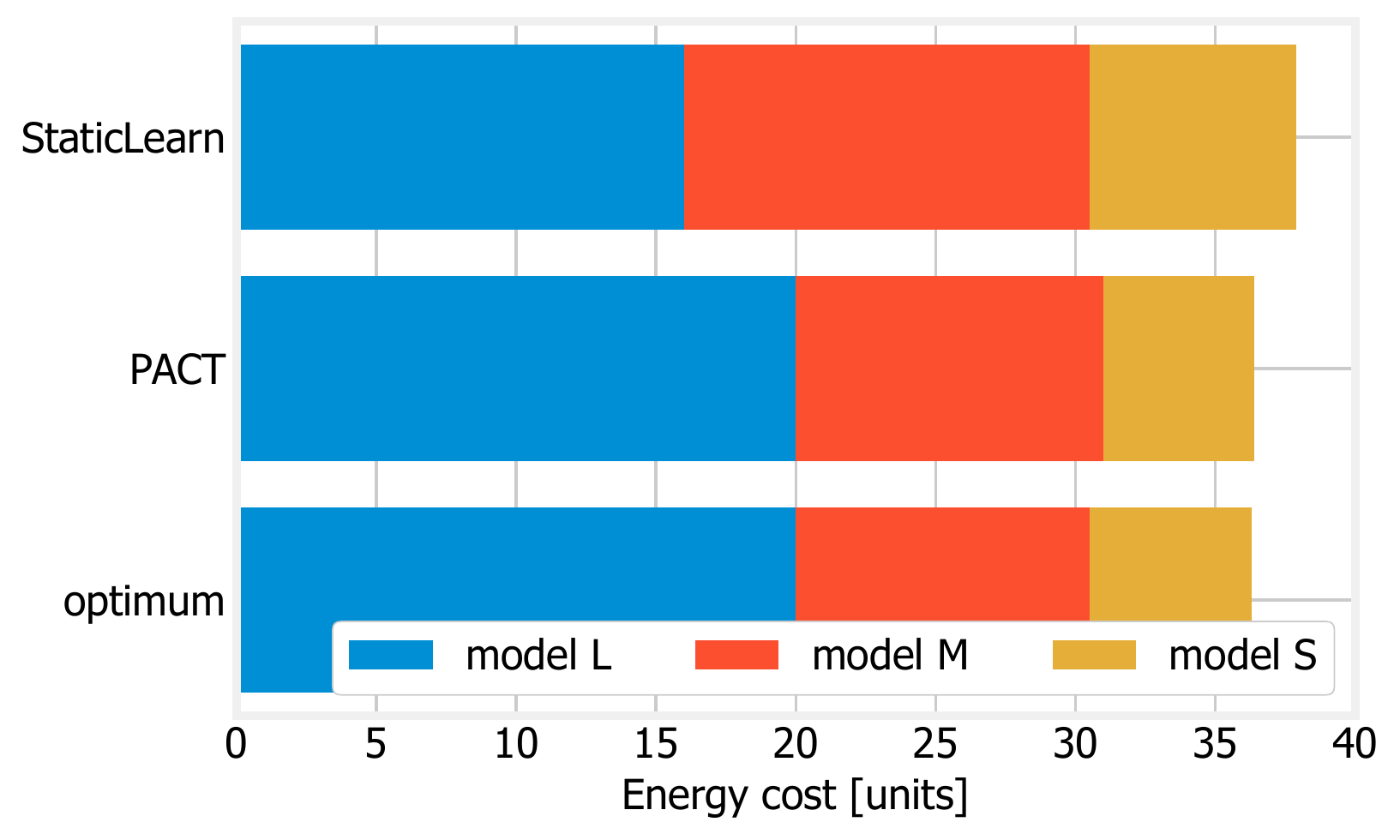}
\includegraphics[width=.32\textwidth]{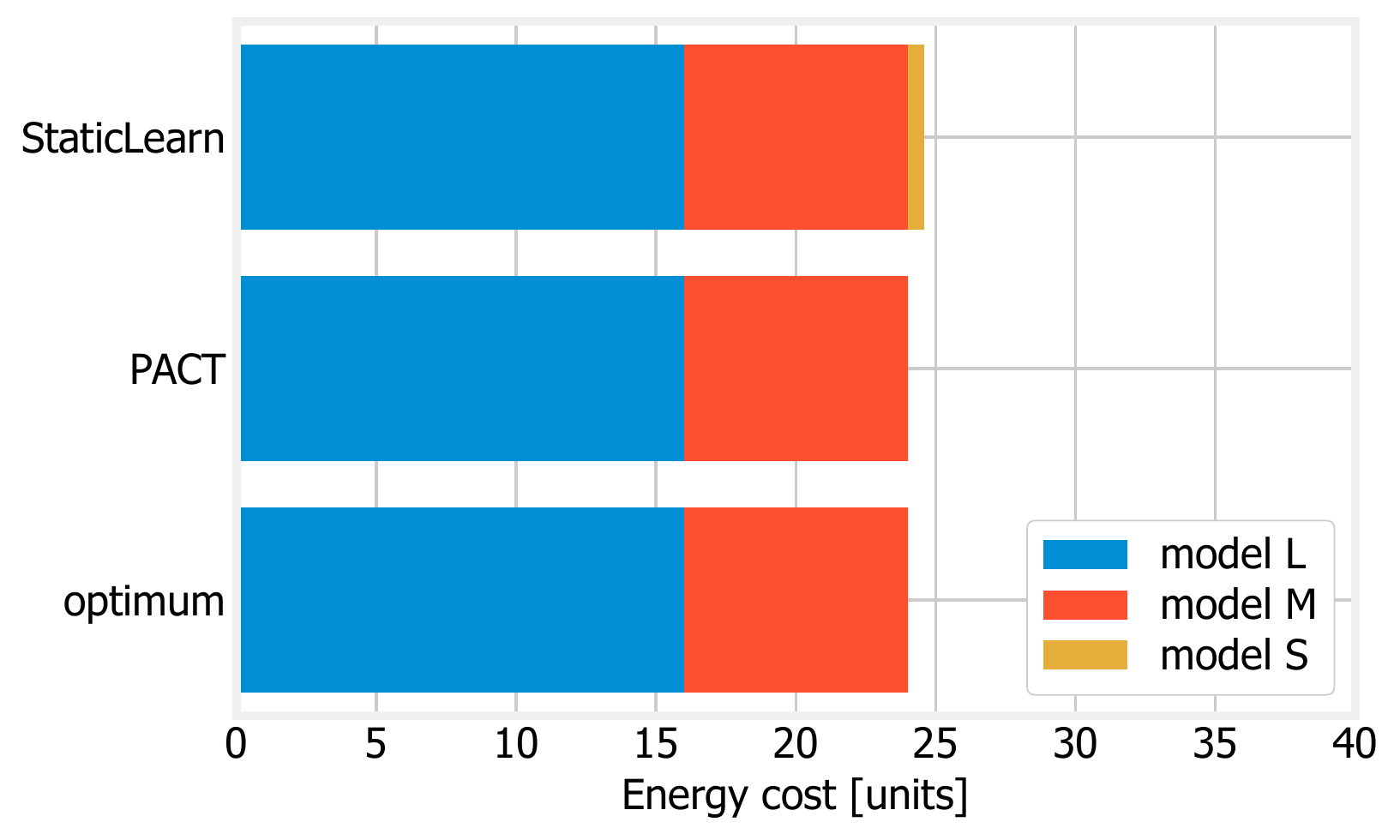}
\includegraphics[width=.32\textwidth]{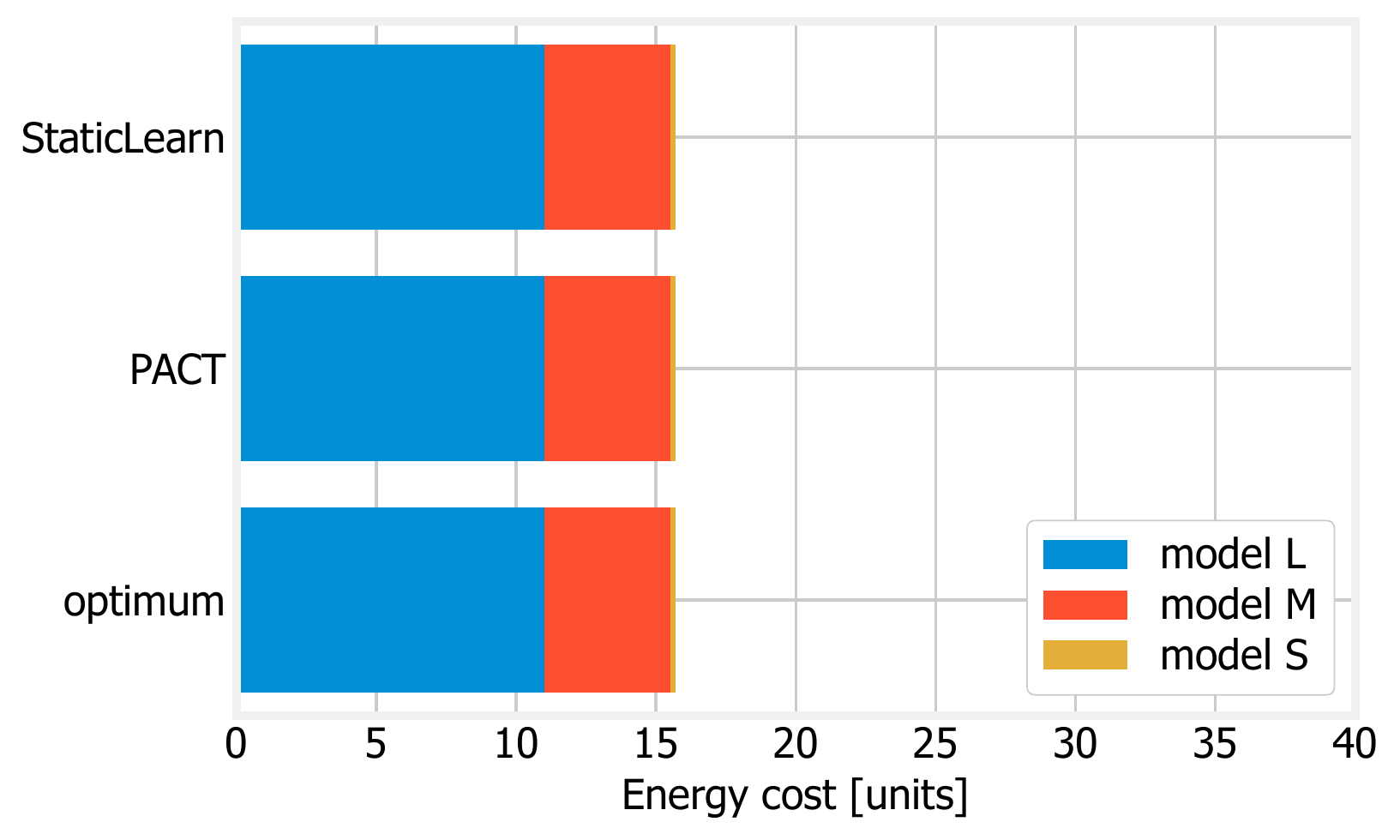}
\caption{
PACT vs. benchmark strategies: energy cost incurred by using different models when $\ell^{\max}=0.15$~(left), $\ell^{\max}=0.3$~(center), $\ell^{\max}=0.45$~(right).
    \label{fig:histos}
} 
\vspace{-4mm}
\end{figure*}

\begin{table}
\centering
\caption{Loss prediction: mean and standard deviation}
\begin{tabular}{|c|c|c|c|}
\hline
Model & MAE  & MIL &  ICP \\
\hline
$\hat{\lambda}^\text{change}$ & $0.13\pm0.01$ & $0.52\pm0.05$  & $0.87\pm0.05$ \\
\hline
$\hat{\lambda}^\text{run}$ & $0.0173\pm 0.0004$ & $0.078\pm0.003$  & $0.90\pm0.01$ \\ \hline
\end{tabular}
\label{tab:results}\vspace{-4mm}
\end{table}

{\bf Benchmark solutions.}
We compare the performance of PACT against the following benchmarks:
(i)
{\em Optimum}: the optimal decisions yielding the minimum cost, found through brute-force search and using the true loss evolution;
(ii) 
 {\em StaticLearn}: model switching is made so as to obtain  similar loss decrease under all three models;
(iii)
{\em OneSwitch}: only two models are used. 
For both the {\em StaticLearn} and {\em OneSwitch} solution, we consider the best decisions they yield for each value of~$\ell^{\max}$. Specifically, for {\em StaticLearn}, we consider the lowest energy cost, feasible strategy obtaining similar (within 5\% margin) loss improvement
from the three models. For {\em OneSwitch}, we consider the lowest energy cost, feasible strategy changing once, considering all combinations of models and changing epochs. Note that most state-of-the-art
works~\cite{gou2021knowledge,gao2021knowru,zhang2021catastrophic} 
 envision pruning once, hence, their performance would be represented by {\em OneSwitch}.

\subsection{PACT performance}\label{sec:res}

First, we evaluate PACT's effectiveness, i.e., how the cost  (i.e., the consumed energy $E(K)$) it yields compares to that of the benchmarks. To this end, \Fig{strats}(left) shows  the cost  associated with each strategy, for different loss targets~$\ell^{\max}$. We can observe that, when~$\ell^{\max}$ is relatively high, all strategies result in very similar performance; on the other hand, they diverge as $\ell^{\max}$~decreases, i.e., as the conditions become more challenging. In particular, PACT outperforms the alternative solutions, and closely matches the optimum, to the point that the corresponding curves almost overlap. Also, only switching models once
has the worst performance, a sign that switching across {\em multiple} models and nodes is indeed beneficial when learning constraints are tight.

\Fig{strats}(center) depicts the time evolution of the loss~$\ell(k)$ for~$\ell^{\max}=0.15$. We can notice 
a power-law behavior -- well captured by our predictors (see \Fig{loss}) and consistent with existing literature~\cite{hestness2017deep,zeulin2021dynamic}, combined with the peaks due to the loss variation
incurred when switching models. Remarkably, PACT makes virtually the same decisions as the optimum, i.e., performs the model switching at (almost) the same times. OneSwitch can only switch once, hence, does so later. As for StaticLearn, in order to achieve similar loss gains under all models, it has to switch from~$L$ to~$M$ earlier than it should, and from~$M$ to~$S$ later, achieving the learning target much later than the alternatives.
Interestingly, PACT achieves the learning target  earlier than the optimum, which would seem counterintuitive until we recall that {\em cost} is the optimization objective \Eq{bellman}, while time is a mere constraint. Accordingly, \Fig{strats}(right) highlights how the optimum indeed takes slightly longer than PACT to reach the objective but  does so at a (marginally) lower cost (see the position of the last marker on the y-axis). This also underlines the importance of making joint decisions about learning and networking aspects, e.g., in this case, to consider both the performance of the models and the cost of the nodes they run on.

\Fig{histos} sheds further light on how different strategies use the network infrastructure. Plots therein show how much energy is spent running each of the three models under the optimum, PACT, and StaticLearn strategies; different plots correspond to different values of~$\ell^{\max}$. Consistently with \Fig{strats}(left), when $\ell^{\max}$~is high or moderate, all strategies make very similar decisions. When $\ell^{\max}$~is low, as in \Fig{histos}(left), the difference between PACT and StaticLearn emerges more clearly;   interestingly, the former spends more energy using model~$L$ and less using model~$S$. Notice that model~$L$ has the {\em highest} cost, hence, one would expect it to be wise to use that for as short as possible. Instead, both PACT and the optimum correctly account for the fact that the quicker learning progress occurring under that model compensates the higher cost it incurs. 
Even more interestingly, when $\ell^{\max}$~is low, PACT and the optimum do not use model~$S$, i.e., they only switch {\em once}. This is consistent with the fact that, as per \Fig{strats}(left), PACT and OneSwitch have the same performance, and highlights the flexibility of PACT in deciding not only {\em when} to switch models, but also on {\em whether} to do so.

\begin{figure*}
\centering
\includegraphics[width=.32\textwidth]{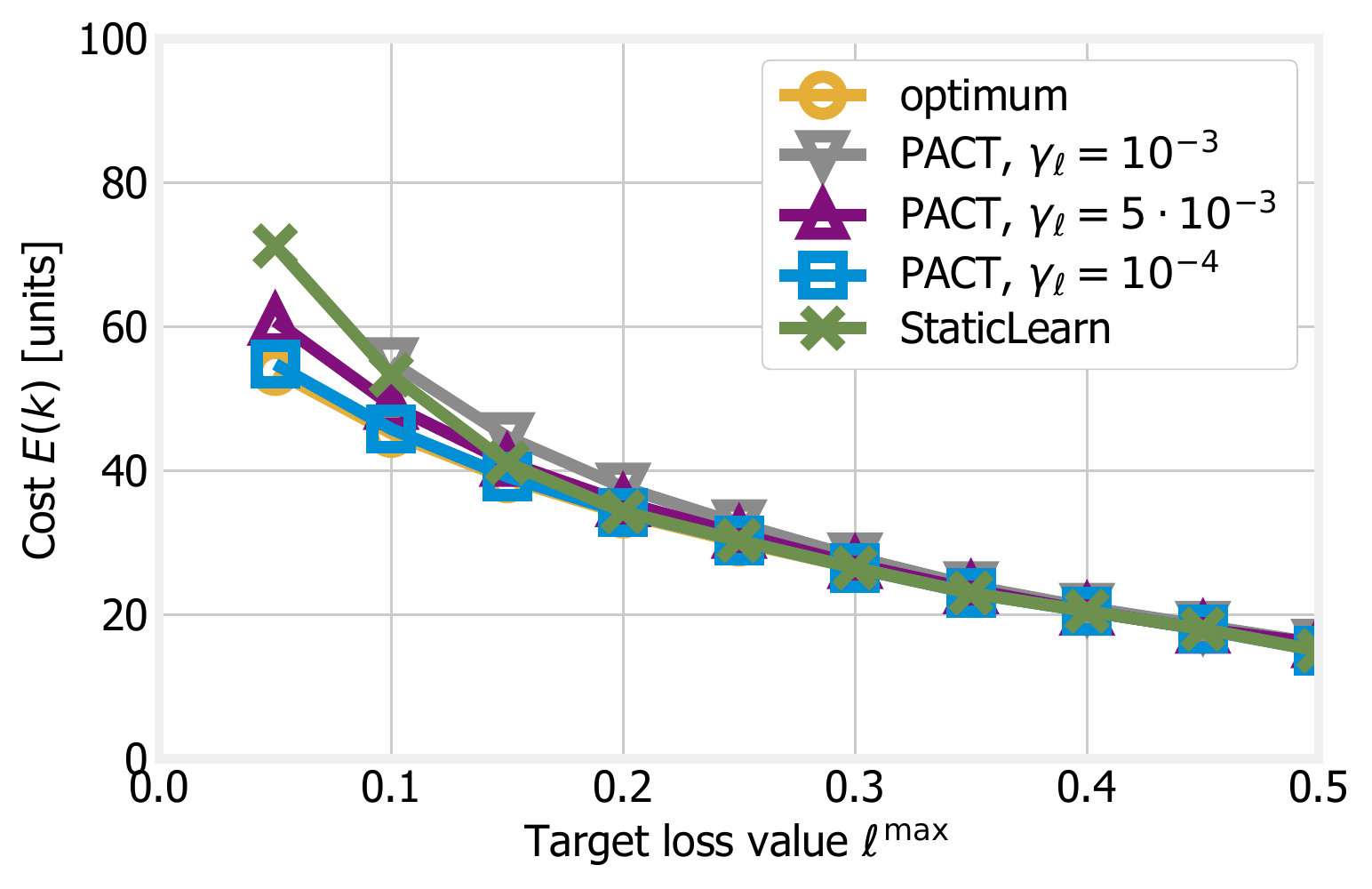}
\includegraphics[width=.32\textwidth]{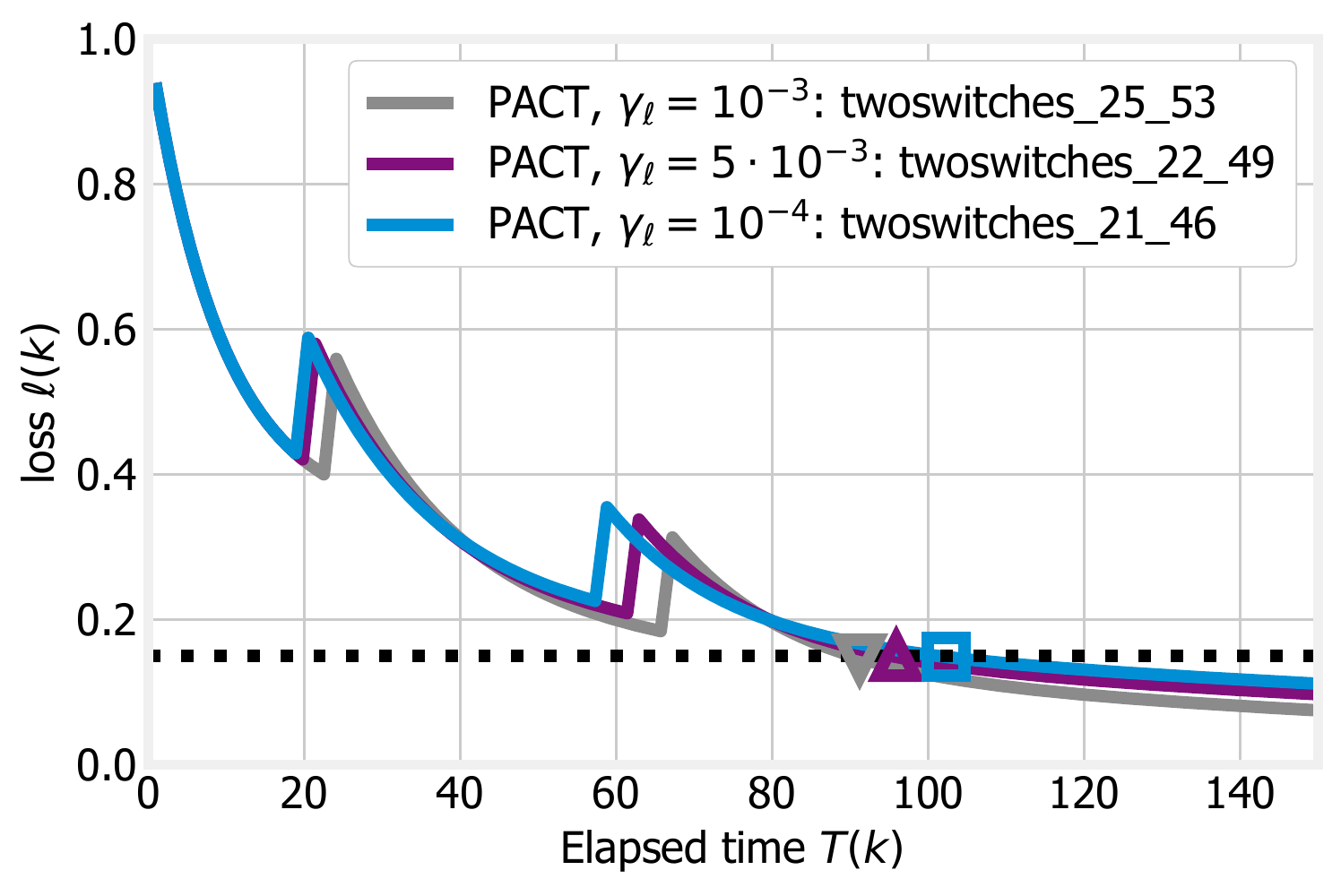}
\includegraphics[width=.32\textwidth]{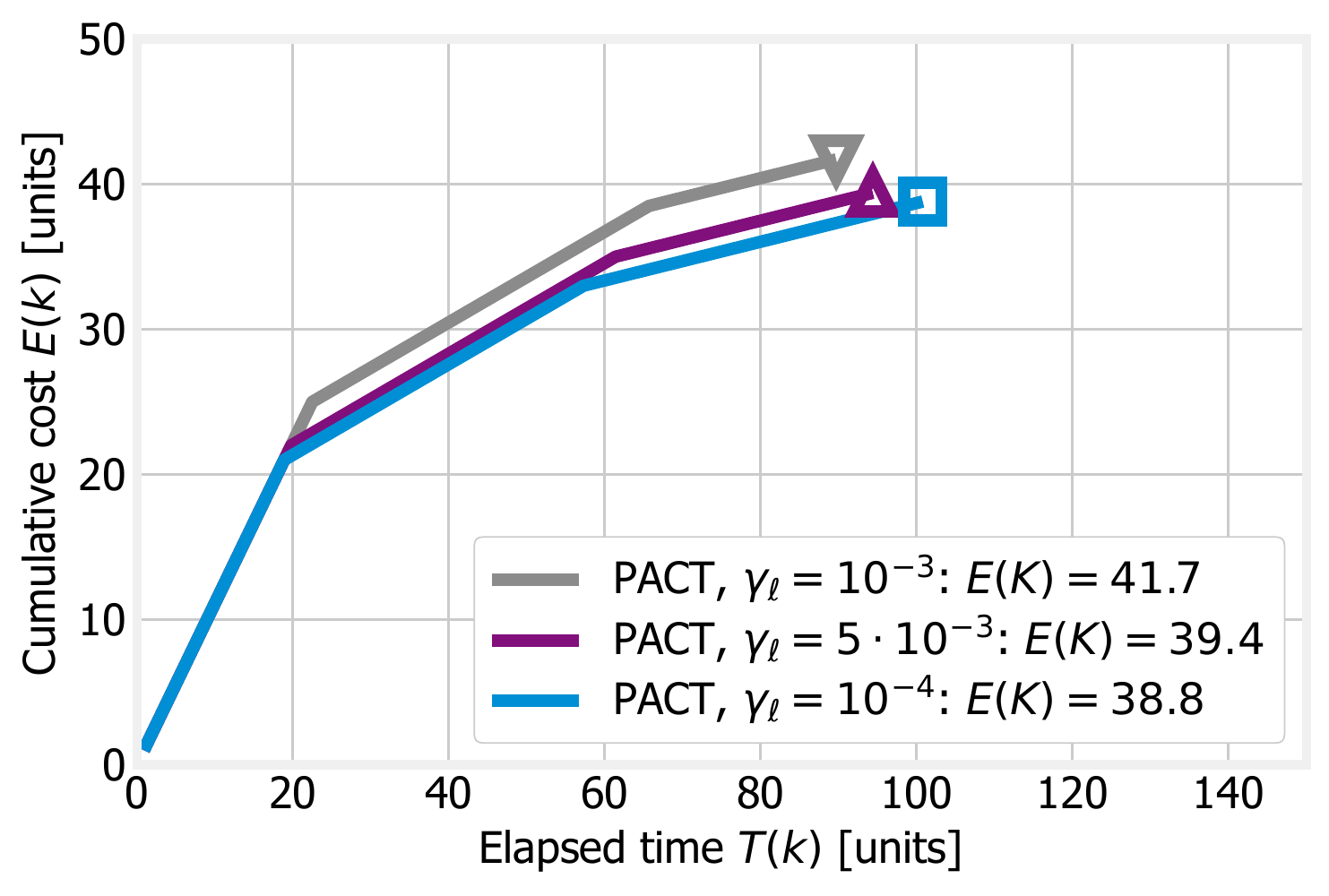}
\caption{
Impact of $\gamma_\ell$ on PACT's performance: energy cost for different values of~$\ell^{\max}$ (left); evolution of the loss (center) and cost (right) when~$\ell^{\max}=0.15$. 
    \label{fig:gamma}
} 
\centering
\includegraphics[width=.32\textwidth]{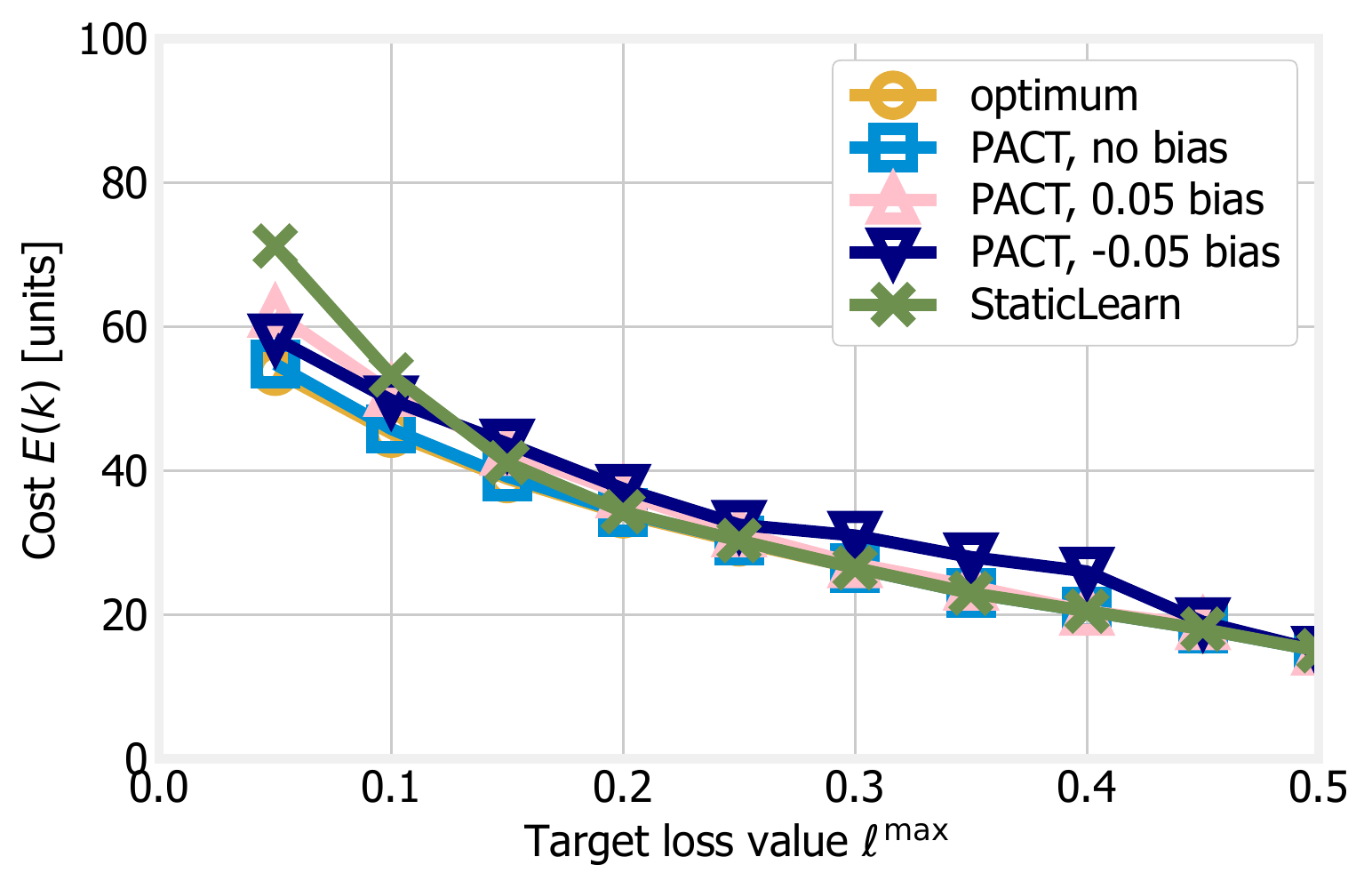}
\includegraphics[width=.32\textwidth]{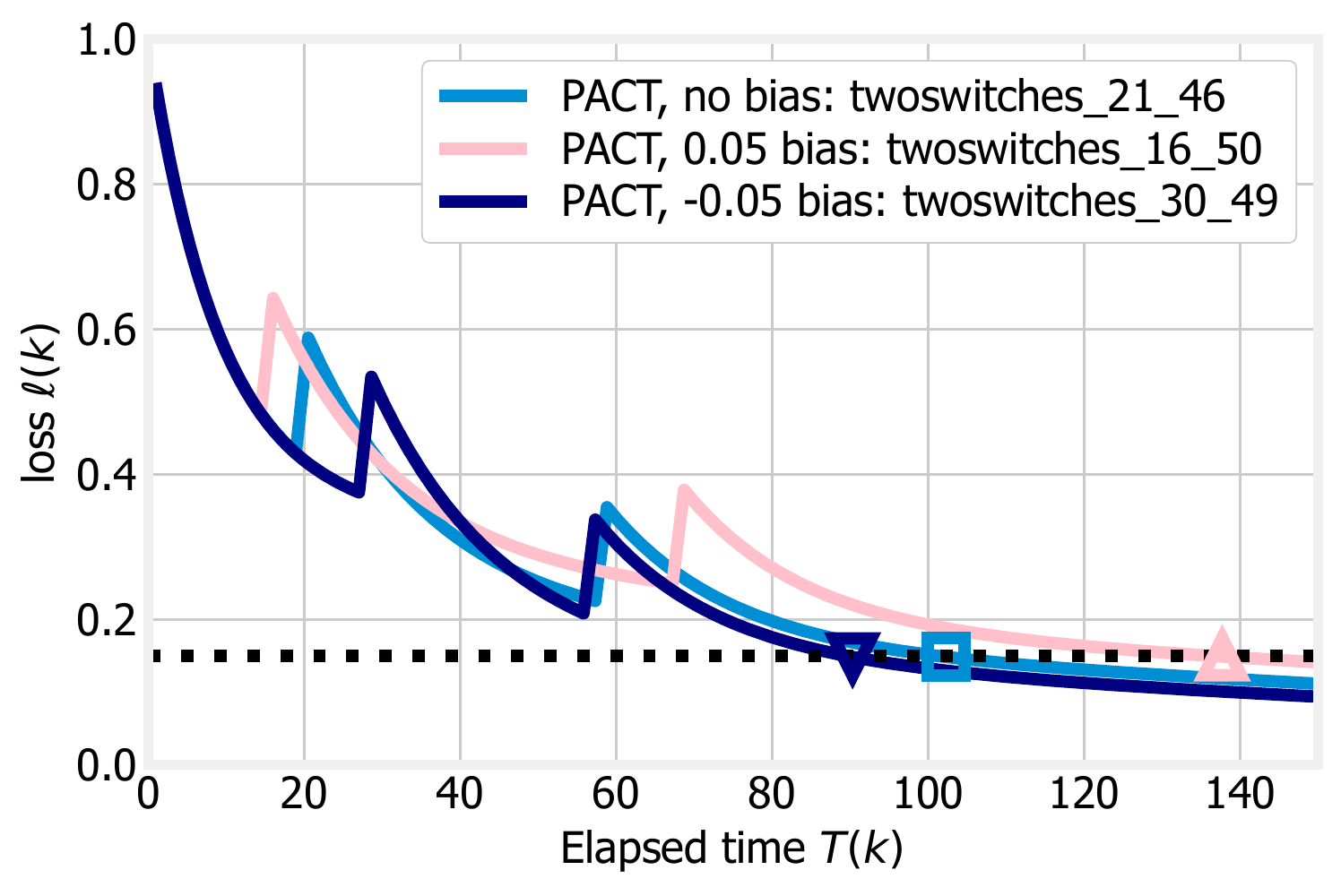}
\includegraphics[width=.32\textwidth]{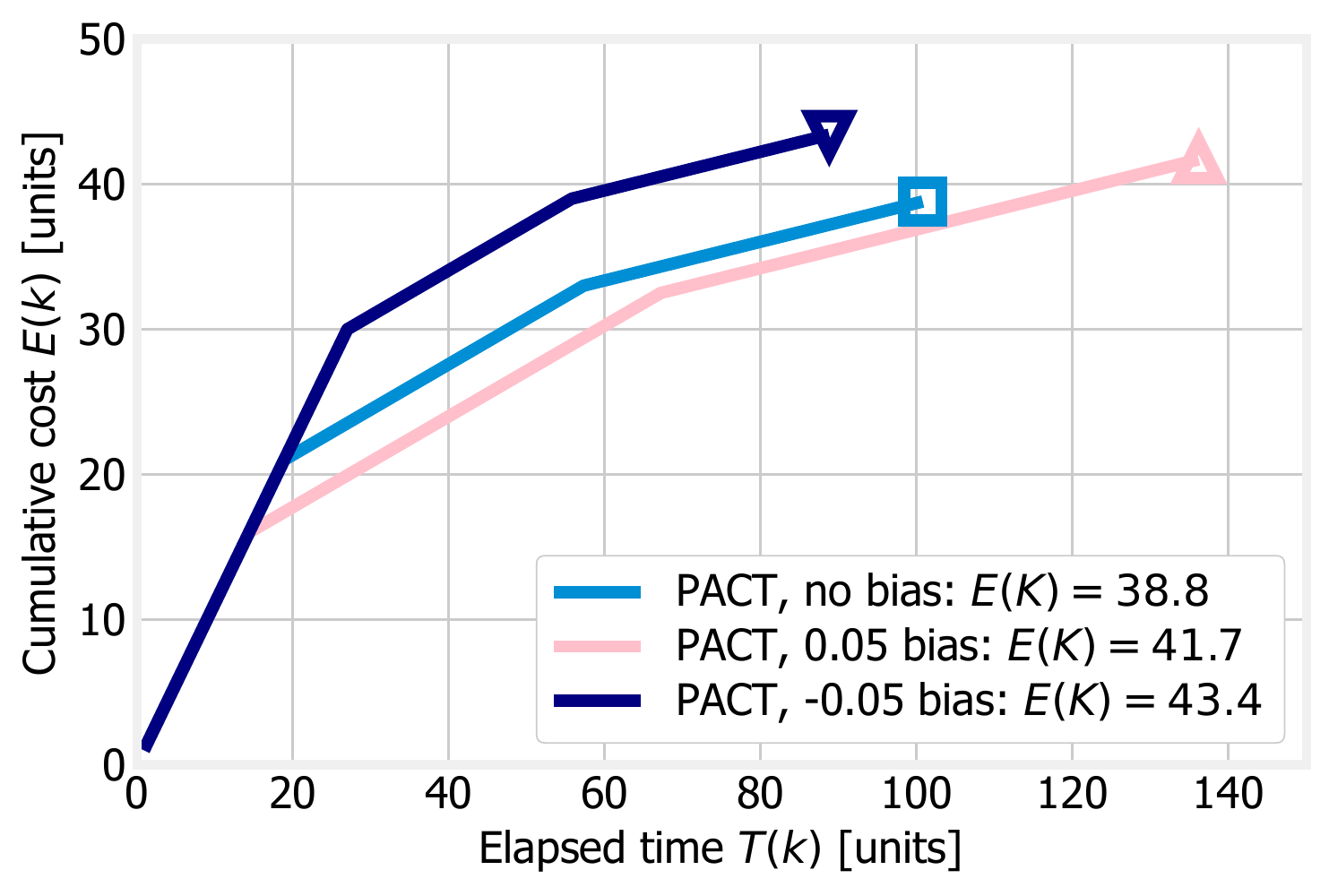}
\caption{
Impact of quality of~$\hat{\lambda}^{\text{run}}$: PACT's energy cost for different values of~$\ell^{\max}$ (left);  loss (center) and cost (right) evolution for~$\ell^{\max}=0.15$. 
    \label{fig:bias}
} 
\vspace{-3mm}
\end{figure*}

Next, we assess the impact of $\gamma_\ell$ and~$\gamma_T$, which control the  
trade-off between PACT's complexity and representation's granularity. 
\Fig{gamma}(left) shows that, while a larger value of $\gamma_\ell$ does indeed decrease PACT's performance, such an effect is limited: even increasing $\gamma_\ell$ by an order of magnitude does not impact PACT's ability to outperform StaticLearn, especially in the most challenging cases when $\ell^{\max}$~is small.

\Fig{gamma}(center), referring to the case~$\ell^{\max}=0.15$, provides some insight on {\em how} a higher~$\gamma_\ell$ affects the decisions made by PACT; specifically, the higher the value of~$\gamma_\ell$, the later switches are made. The reason lies in \Line{mkg-fix-l1} of \Alg{makegraph}, and more exactly in the ceiling operator therein. Increasing~$\gamma_\ell$ leads to overestimating the loss resulting from a particular action, hence, to assume that further gains could be made under the current model, while that is not the case. For the same value of~$\ell^{\max}$, \Fig{gamma}(right) highlights how these later switches result in a higher cost -- though, similarly to \Fig{strats}(right), {\em not} necessarily in a longer learning time.

Finally, 
we further assess how well PACT can deal with loss estimation errors, by adding a {\em bias} to the prediction output for model~$L$. 
\Fig{bias}(left) shows that  positive and negative biases yield  similar performance decrease. Also, except for the simple cases when $\ell^{\max}$~is very large, PACT  outperforms StaticLearn even in the presence of a bias.
\Fig{bias}(center), referring to the case~$\ell^{\max}=0.15$, shows how biases on the loss variations prediction influence PACT's decisions. Consistently with \Fig{gamma}(center), underestimating $L$'s performance leads to a later switch, while overestimating it has the opposite effect. It is also worth noting the times of the {\em second} switch, from~$M$ to~$S$: PACT can leverage its bias-free knowledge of the performance of~$M$ and~$S$, compensating for the misguided decisions it made earlier. \Fig{bias}(right) underlines, similarly to \Fig{gamma}(right), that switching earlier results in a longer training time, though this does not necessarily result in a higher energy cost. 

\section{Related Work
\label{sec:relwork}}


\noindent
{\em Model switching and compression.}
The two most popular techniques for model compression are KD and pruning. 
In KD~\cite{gou2021knowledge}, 
a small-size (student) model does not learn directly from data, but mimics the behavior of the large (teacher) model. 
Studies focus on task generalization~\cite{gao2021knowru}, 
and data heterogeneity \cite{zhang2021catastrophic}.
As for pruning, a very  effective technique is 
{\em structured}  pruning~\cite{wen2016learning}, which removes whole parts of a DN (e.g.,  rows or columns of the parameter matrix).
Finally, recent work~\cite{he2018amc} proposes the use of RL to control pruning.

\noindent
{\em Hybrid approaches.}
Some  works explore how to alternate distributed learning schemes  such as Split Learning (SL), FL, and KD. An example is~\cite{matsubara2020head}, which splits the DNN architecture into head and tail, and   replaces the former with its distilled version. 
\cite{zhou2021communication} seeks to reduce the network delay incurred by FL by performing communication and local learning concurrently.  
In a similar setting, \cite{chopra2021adasplit} optimizes the computation, communication, and cooperation aspects of FL in resource-constrained scenarios. \cite{sen2022data} leverages RL to identify the best split of a learning task 
across the available network nodes. 
\cite{marfoq2022personalized} targets highly heterogeneous scenarios by 
proposing a {\em personalized learning} where a different model is trained at each device.

\noindent
{\em Resource-aware distributed ML.} 
In the context of FL, several works focus on {\em selecting} the participating nodes,  accounting for their speed~\cite{wang2019adaptive,abdelmoniem2021resource}, quantity~\cite{marfoq2022personalized,abdelmoniem2021resource,marfoq2022personalized,malandrino2021federated} and quality~\cite{malandrino2021federated,wu2021fast} of local data, the speed and reliability of their network~\cite{wang2019adaptive,zhou2021communication} as well as trust~\cite{imteaj2020fedar}. The basic trade-off balances the need to learn more during each epoch 
with the need to shorten the duration of epochs.   
Other works~\cite{zaw2021energy,malandrino2022energy} target a more general scenario, where  DNN layers  can be run, and possibly be duplicated, at different nodes. This requires balancing the opportunity to use fast learning nodes with the network delays resulting from moving data between nodes. 
Interestingly, recent work (e.g., \cite{tan2019efficientnet}) has aimed at creating energy-efficient DNN architectures, offering better trade-offs between energy efficiency and learning effectiveness.  

\noindent
{\em Distributed learning characterization.} 
Early work~\cite{neglia} studies the convergence of distributed learning, identifying the latent trade-off between involving more nodes 
and exploiting fewer, faster nodes. 
The experiments in~\cite{hestness2017deep} report a power-law behavior, with the exponent depending on the quantity of data, and the model architecture shifting the error, but not reducing the exponent itself. 
Other works focus on FL and derive exponential bounds on the loss~\cite{zeulin2021dynamic,li2019convergence}.
%
%
Studies focusing on KD are more rare. Examples include~\cite{phuong2019towards}, which models the teacher-to-student translation as a price to pay on the loss, and \cite{rahbar2020unreasonable} that provides a per-iteration characterization of KD. 

\section{Conclusions\label{sec:concl}}
We addressed the problem of matching the  training and compression  of DNN models, with the aim to minimize  energy consumption while meeting  learning performance and system constraints. To do so,  we used  approximate dynamic programming and developed the PACT algorithmic framework to overcome the problem's NP-hardness. PACT uses a time-expanded graph to model the system  and leverages both a data-driven and a theoretical approach for predicting the loss behavior as training decisions are made. Results show that PACT matches the minimum energy consumption very closely (with worst-case polynomial complexity),  while meeting the learning quality and time requirements.

\section*{Acknowledgement}

This work was supported by the European Union's Horizon Europe program through the project CENTRIC under grant agreement No. XXXXXXX.

\bibliographystyle{IEEEtran}
\bibliography{refs}

\end{document}